%% file: main.tex
\documentclass[10pt,twocolumn,letterpaper]{article}

\usepackage{cvpr}              

\usepackage[accsupp]{axessibility}  

\usepackage{graphicx}
\usepackage{amsmath}
\usepackage{amsthm}
\usepackage{float}
\usepackage{amssymb}
\usepackage{booktabs}
\usepackage{bm}
\usepackage{multirow}
\usepackage[table]{xcolor}
\usepackage{tabularx}
\usepackage{epsfig}
\usepackage[outdir=./]{epstopdf}
\usepackage{wrapfig}

\usepackage{url} 

\usepackage{nicematrix}

\usepackage[pagebackref,breaklinks,colorlinks]{hyperref}

\usepackage{algorithm}
\usepackage{algpseudocode}

\newcommand{\greenrp}[1]{\textcolor[RGB]{129, 211, 26}{#1}} 
\newcommand{\bluerp}[1]{\textcolor[RGB]{0, 176, 240}{#1}} 
\newcommand{\orangerp}[1]{\textcolor[RGB]{255, 147, 0}{#1}} 

\usepackage[capitalize]{cleveref}
\crefname{section}{Sec.}{Secs.}
\Crefname{section}{Section}{Sections}
\Crefname{table}{Table}{Tables}
\crefname{table}{Tab.}{Tabs.}

\definecolor{myred}{RGB}{214, 39, 40}

\def\cvprPaperID{1495} 
\def\confName{CVPR}
\def\confYear{2023}

\newcommand{\ouralg}{DSI}

\definecolor{MUTEDLIGHTGREEN}{RGB}{219, 245, 206}
\newcommand{\g}{{\cellcolor{MUTEDLIGHTGREEN}}}
\renewcommand{\u}[1]{{\cellcolor{MUTEDLIGHTGREEN}}\textbf{#1}}

\input{cmd}

\newcommand{\mycolorbox}[1]{%
  \begingroup\setlength{\fboxsep}{1pt}%
  \colorbox{MUTEDLIGHTGREEN}{\vphantom{Ay}#1}%
  \endgroup
}

\def\thanks#1{\protected@xdef\@thanks{\@thanks
\protect\footnotetext{#1}}}
\makeatother

\begin{document}


\input{maintext}

\input{appendix}

\clearpage
{\small
\bibliographystyle{ieee_fullname}
\bibliography{bib}
}


\end{document}

%% file: cmd.tex
\usepackage{amsmath,amsfonts,bm,amssymb,mathtools,amsthm}
\usepackage{color,xcolor}

\usepackage{hyperref}       %
\hypersetup{
  colorlinks,
  linkcolor={red!50!black},
  citecolor={blue!50!black},
  urlcolor={blue!80!black}
  }
  \definecolor{orange}{HTML}{ff7f0e}
  \definecolor{blue}{HTML}{1f77b4}
  
\usepackage{url}

\usepackage{algorithm}
\usepackage{algpseudocode}
\usepackage{algorithmicx}

\usepackage{microtype}
\usepackage{lipsum}
\usepackage{graphicx}
\usepackage{subcaption}
\usepackage{latexsym}
\usepackage{sidecap}
\usepackage{caption}
\usepackage{booktabs} %
\usepackage{amsfonts}       %
\usepackage{nicefrac}       %
\usepackage{comment}
\usepackage{enumitem}
\usepackage{wrapfig,tikz}
\usepackage{longtable,fancyhdr}
\usepackage{adjustbox, bigstrut, tabularx, multirow, makecell, diagbox}
\usepackage{graphicx,float}
\usepackage{stackrel}
\usepackage{color}
\usepackage{colortbl}
\usepackage{theoremref}
\usepackage{cases}
\usepackage{stmaryrd}
\usepackage{mathabx}
\usepackage{dsfont}
\usepackage{arydshln}

\makeatletter
\usepackage{xspace} 
\def\@onedot{\ifx\@let@token.\else.\null\fi\xspace}
\DeclareRobustCommand\onedot{\futurelet\@let@token\@onedot}

\definecolor{blue1}{RGB}{0,128,255}
\definecolor{blue3}{RGB}{0,0,128}
\definecolor{darkpastelgreen}{rgb}{0.01, 0.75, 0.24}
\definecolor{cerulean}{rgb}{0.0, 0.48, 0.65}

\newcommand{\mbb}[1]{\mathbb{#1}}
\newcommand{\mcal}[1]{\mathcal{#1}}

\def\vs{\emph{vs}\onedot}

\definecolor{darkgreen}{rgb}{0,0.6,0}

\newtheorem{theorem}{Theorem}

\def\eqref#1{equation~\ref{#1}}

\def\1{\bm{1}}

\def\vtheta{{\bm{\theta}}}

\def\vs{{\bm{s}}}

\DeclareMathAlphabet{\mathsfit}{\encodingdefault}{\sfdefault}{m}{sl}
\SetMathAlphabet{\mathsfit}{bold}{\encodingdefault}{\sfdefault}{bx}{n}

\newcommand{\E}{\mathbb{E}}

\newcommand{\ud}{\mathop{}\!\mathrm{d}}

\newcommand{\norm}[1]{\left\lVert#1\right\rVert}

%% file: maintext.tex
\title{
Distribution Shift Inversion for Out-of-Distribution Prediction
}

\author{\bf Runpeng Yu \quad 
Songhua Liu \quad
Xingyi Yang \quad
Xinchao Wang$^{\dagger}$ \thanks{ $^{\dagger}$ Corresponding author.}\\
National University of Singapore\\
{\tt\small \{r.yu,songhua.liu,xyang\}@u.nus.edu} \quad {\tt\small xinchao@nus.edu.sg}
}

\twocolumn[{
\renewcommand\twocolumn[1][]{#1}%
\maketitle
\vspace{-4em}
\begin{center}
    \centering
    \captionsetup{type=figure}
    \includegraphics[width=\linewidth]{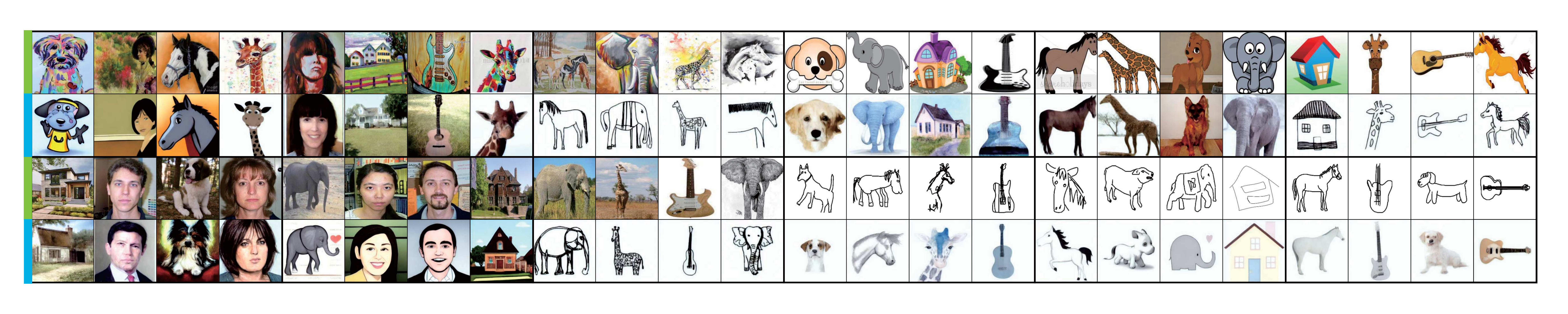}
    \vspace{-2em}
    \captionof{figure}{Transformed OoD samples from PACS dataset. {\bf \greenrp{Odd rows}} show the original OoD images, and {\bf \bluerp{even rows}} show their transformation results to the source distribution,
    obtained by the proposed \ouralg.
    Please zoom in for better visualization. }
  \label{fig:samples}
\end{center}
}]

{
  \renewcommand{\thefootnote}{\fnsymbol{footnote}}
  \footnotetext[2]{Corresponding author.}
}

\begin{abstract}
    \vspace{-1em}
    Machine learning society has witnessed the emergence of a myriad of Out-of-Distribution (OoD) algorithms, which address the distribution shift between the training and the testing distribution by searching for a unified predictor or invariant feature representation. 
    However, the task of directly mitigating the distribution shift in the unseen testing set is rarely investigated, due to the unavailability of the testing distribution during the training phase and thus the impossibility of training a distribution translator mapping between the training and testing distribution. 
    In this paper, we explore how to bypass the requirement of testing distribution for distribution translator training and make the distribution translation useful for OoD prediction.
    We propose a portable Distribution Shift Inversion (\ouralg) algorithm, in which, before being fed into the prediction model, the OoD testing samples are first linearly combined with additional Gaussian noise and then transferred back towards the training distribution using a diffusion model trained only on the source distribution.
    Theoretical analysis reveals the feasibility of our method.
    Experimental results, on both multiple-domain generalization datasets and single-domain generalization datasets, show that our method provides a general performance gain when plugged into a wide range of commonly used OoD algorithms. Our code is available at \href{https://github.com/yu-rp/Distribution-Shift-Iverson}{https://github.com/yu-rp/Distribution-Shift-Iverson}.
\end{abstract}


\section{Introduction}
The ubiquity of the distribution shift between the training and testing data in the real-world application of machine learning systems induces the study of Out-of-Distribution (OoD) generalization (or domain generalization). ~\cite{ood-bench,domain-bed,ood-survey,ood-survey2} Within the scope of OoD generalization, machine learning algorithms are required to generalize from the seen training domain to the unseen testing domain without the independent and identically distributed assumption.
The bulk of the OoD algorithms in previous literature focuses on promoting the generalization capability of the machine learning models themselves by utilizing the domain invariant feature~\cite{irm,cdan,dann}, context-based data augmentation~\cite{sagnet, mixup}, distributionally robust optimization~\cite{groupDRO}, subnetwork searching~\cite{subnet}, neural network calibration~\cite{calibration}, etc. 

In this work, orthogonal to enhancing the generalization capability of the model, we consider a novel pathway to OoD prediction. On the way, the testing(target) distribution is explicitly transformed towards the training(source) distribution to straightforwardly mitigate the distribution shift between the testing and the training distribution. 
Therein, the OoD prediction can be regarded as a two-step procedure, (1) transferring testing samples back towards training distribution, and (2) drawing prediction. The second step can be implemented by any OoD prediction algorithm. In this paper, we concentrate on the exploration of the first step, the distribution transformation.

Unlike previous works on distribution translation and domain transformation, in which certain target distribution is accessible during the training phase, here the target distribution is arbitrary and unavailable during the training.
We term this new task as Unseen Distribution Transformation (UDT), in which a domain translator is trained on the source distribution and works to transform unseen target distribution towards the source distribution. The uniqueness, as well as the superiority, of UDT is listed as followings. 
\begin{itemize}[itemsep=2pt,topsep=0pt,parsep=0pt]
    \item UDT puts no requirements on the data from both source and testing distribution like previous works do. This is practically valuable, because the real-world testing distributions are uncountable and dynamically changing.
    \item UDT is able to transform various distributions using only one model. However, the previous distribution translator works for the translation between certain source and target distributions. With a different source-target pair, a new translator is required.
    \item Considering the application of UDT in OoD prediction, it is free from the extra assumptions commonly used by the OoD generalization algorithms, such as the multi-training domain assumption and the various forms of the domain invariant assumption.
\end{itemize}

Despite the advantages, the unavailability of the testing distribution poses new difficulty. Releasing this constraint, the idea of distribution alignment is well established in domain adaptation (DA). Wherein, a distribution translator 
is trained with the (pixel, feature, and semantic level) cycle consistency loss~\cite{i2ida,cycada,segmentationda}. However, the training of such distribution transfer modules necessitates
the testing distribution,
which is unsuitable under the setting of OoD prediction and makes the transplant of the methods in DA to OoD even impossible.

\input{2figs/toy_example}

To circumvent the requirement of testing distribution during training time, we propose a novel method, named Distribution Shift Inversion  (\ouralg). 
Instead of using a model transferring from testing distribution to training distribution, an unconditional generative model, trained only on the source distribution, is used, which transfers data from a reference noise distribution to the source distribution. 
The method operates in two successive parts. First, the OoD target distribution is transferred to the neighborhood of the noise distribution and aligned with the input of the generative model, thereafter we refer to this process as the forward transformation. 
The crux of this step is designing to what degree the target distribution is aligned to the noise distribution. 
In our implementation, the forward transformation is conducted by linearly combining the OoD samples and random noise with the weights controlling the alignment.
Then, in the second step, the outcome of the first step is transferred towards the source distribution by a generative model, thereafter we refer to this process as the backward transformation. 
In this paper, the generative model is chosen to be the diffusion model~\cite{ddpm}.
The superiority of the diffusion model is that its input is the linear combination of the source sample and the noise with varying magnitude, which is in accord with our design of the forward transformation and naturally allows strength control. Comparatively, VAE~\cite{vae} and GAN~\cite{gan} have a fixed level of noise in their input, which makes the forward transformation strength control indirect. 
Our theoretical analyses of the diffusion model also show the feasibility of using the diffusion model for UDT. 

\textbf{Illustrative Example.} A one-dimensional example is shown in \cref{fig:toy}.
The example considers a binary classification problem, in which, given label, the conditional distributions of the samples are Gaussian in both the source and the testing domain. The testing distribution is constructed to be OoD and located in the region where the source distribution has a low 
density. The diffusion model is trained only on the source distribution. Passing through the noise space alignment and diffusion model transformation, the OoD samples are transformed to the source distribution with limited failure of label preservation. 

\textbf{Transformed Images.} \cref{fig:samples} shows some transformation results of OoD images towards the source distribution. The observation is twofold. (1) The distribution (here is the style) of the images is successfully transformed. All of the transferred images can be correctly classified by the ERM model trained on the source domain. (2) The transformed images are correlated to the original images. Some structural and color characteristics are mutually shared between them. This indicates that the diffusion model has extracted some low-level information and is capable to preserve it during the transformation. We would like to highlight again that, during the training, the diffusion model is isolated from the testing domain.

Our contributions are therefore summarized as:
\begin{itemize}[itemsep=2pt,topsep=0pt,parsep=0pt]
    \item We put forward the unseen distribution transformation (UDT) and study its application to OoD prediction.
    \item We offer theoretical analyses of the feasibility of UDT. 
    \item we propose \ouralg, a sample adaptive distribution transformation algorithm for efficient distribution adaptation and semantic information preservation.
    \item We perform extensive experiments to demonstrate that our method is suitable for various OoD algorithms to achieve performance gain on diversified OoD benchmarks. On average, adding in our method produces 2.26\% accuracy gain
    on multi-training domain generalization datasets and 2.28\% on single-training domain generalization datasets.
\end{itemize}

\section{Related Work}\label{sec:related_work}
\subsection{Out-of-Distribution Generalization}
To achieve OoD generalization, diverse methods have been proposed. 

Causal inference methods extract the invariant feature among training domains and build up a unified predictor for all domains. 
For example,
CNBB~\cite{cnbb} down-weights the samples which introduce big changes to the feature space by evaluating the causal effect of each sample; 
ICP~\cite{icp} find the subset of features satisfying the property that the conditional probability of the target given this set of features is invariant among the domains; 
IRM~\cite{irm} regularizes the ERM loss with the norm of the gradient of loss corresponding to the latent features;  IB-IRM~\cite{ibirm} extends the IRM to the case, where the invariant features are only partially informative, by adding an extra entropy loss term of the latent feature.

Context-based data augmentation methods enrich the training feature space by linear interpolation between the features of inter-domain and intra-domain samples~\cite{mixup}; or combining the normalized feature of one sample with the mean and variance of the feature of another sample~\cite{sagnet}. 

Distributional Robust Optimization method, like GroupDRO~\cite{groupDRO}, theoretically formulates the OoD generalization as a min-max optimization and practically up-weights the domains with large losses in an online learning style to guarantee the model fit on all domains. 

Feature alignment methods learn a unified feature representation among all training domains with the assumption that this representation is shared by the testing domain data. For example, 
DANN~\cite{dann} and CDANN~\cite{cdan} align the marginal and conditional feature distribution by domain adversarial training, respectively; 
CORAL~\cite{coral} matches the covariance of the features in every training domain;
MASF~\cite{masf} proposes model-agnostic episodic learning to regularize the semantic structure of the feature space;
using the contrastive learning as the regularization term, SelfReg~\cite{selfreg} minimizes the distance between the features of the samples within the sample class; CAD and ConCAD~\cite{cad} 
minimize the mutual information between the feature and domain variable as well as the negative mutual information between the feature of original samples and the feature of augmented samples.

Context feature separation methods recognize the context feature by an extra context discriminator and disentangle them from the category feature by orthogonality penalty~\cite{decaug}.

Gradient manipulation methods are designed based on the intuitions that (1) to train a unified predictor, only the gradient common across all domains should be used; or the intuition that (2) the spurious correlation, which leads to larger gradients, is easier to be fitted to and prevent the algorithm from learning other features.
For example, 
IGA~\cite{iga} and SD~\cite{sd} penalize the variance of the gradients and the $l_2$ norm of the logits, respectively;
RSC~\cite{rsc} only takes the samples whose gradients are small into consideration;
MLDG~\cite{mldg} updates the parameters only when there are performance gains in two separated parts of the training dataset;
ANDMask~\cite{andmask} and its ReLU-smoothed version SANDMask~\cite{sandmask} only update the parameter whose gradients in most of the domains have the same sign.

Different from the previous works, \ouralg\  tackles the OoD generalization problem by test time sample adaption. Instead of increasing the OoD generalization ability of the model, \ouralg\  aims to mitigate the distribution shift of the testing samples by shifting them back towards the source distribution. Though input enhancing is investigated in dataset distillation~\cite{dc} and domain adaptation~\cite{i2ida,cycada,segmentationda}, we are the first to test this idea in OoD generalization. Besides its novelty, \ouralg\  is orthogonal to the previous algorithms, which allows it to be portably used as a common technique for OoD tasks.

\subsection{Diffusion Model}
Neural network reuse has drawn recent interest~\cite{nreuse1,nreuse2,nreuse3,nreuse4}. In this work, we consider reusing pre-trained diffusion models.
With promising performance and theoretical explanation, the diffusion model has been intensively investigated in the field of time series generalization/imputation/prediction~\cite{timegen,timeimp,timepred,voicegen}, image generalization/editing/inpainting~\cite{imgedit1,imgedit2,imggen1,imggen2,imggen3,imggen4,imggen5,imginpt}, text generalization/modeling~\cite{textgen,textmodeling}, text2image generalization~\cite{text2imge1,text2imge2,text2imge3,text2imge4}, text2speech generalization~\cite{text2speech}, video generalization~\cite{videogen}, graph generalization~\cite{graphgen1,graphgen2}, 3D point cloud generalization~\cite{pointcloudgen1,pointcloudgen2,pointcloudgen3}. Besides, these generation tasks, diffusion model conduces to the downstream tasks such as image segmenting~\cite{dmsag}, testing time sample adaptation for corrupted images~\cite{b2s}, and adversarial attack defense~\cite{dm4ad}.
In \ouralg, we explore a new application of the diffusion model in which it enhances the OoD prediction by transferring the OoD samples towards the source distribution. 

\subsection{Prediction Confidence Estimation}
In \ouralg, the prediction confidence score is used for selecting poorly predicted samples and achieving an adaptive distribution shift inversion at the sample level.  By satisfying the partial order relationship of uncertainty, commonly used loss functions, such as softmax cross entropy, are suitable confidence metrics.~\cite{commonloss} Other softmax-based scores, such as the maximum class probability~\cite{confidence_score_baseline, selective_classification}, true class probability~\cite{tcp} and the KL-divergence between the softmax distribution and the uniform distribution~\cite{confidence_score_baseline} are also widely used. Based on the Bayesian method, Monte Carlo dropout~\cite{mcdropout}, use the standard uncertainty criteria (e.g., variance, entropy) of the stochastic network predictions as the confidence score. Taking a modified nearest-neighbor classifier as the reference, trust score~\cite{knnscore} uses a minimal distance ratio as the confidence score. We note that measurements in the field of OoD detection~\cite{typicality,oodd1,complexity,llr,detectenergy} and sample difficulty estimation may also serve as proper filtering metrics for \ouralg. But, in this paper, we prefer to establish a new OoD prediction framework and leave the discovery of the potential filtering metrics as future work.

\section{Preliminary}
\label{sec:prelim}

The task of sampling an instance $\bm{x}$ of a random variable $\bm{X}\in\mathcal{R}^d$ with distribution $p(\bm{x})$ is generally intractable due to the fact that $p(\bm{x})$ is complex or unknown. Diffusion models, such as ScoreFlow~\cite{scoreflow} and DDPM~\cite{ddpm}, first samples from a simple distribution $\rho(\bm{x})$ and then iteratively transform sample $\bm{x}$ to make its distribution consistent with $p(\bm{x})$. 

Under the continuous time setting, diffusion model formulates the forward transformation from $p(\bm{x})$ to $\rho(\bm{x})$ as a stochastic process $\{x_t\}_{t=0}^T$ following the Stochastic Differential Equation (SDE)
\begin{align} \label{eq:prelim:forward_sde}
    d\bm{x} = f(\bm{x},t)dt + g(t)d\bm{w},
\end{align}
where $f(\cdot,t):\mathcal{R}^d\rightarrow\mathcal{R}^d$ is named drift coefficient, $g(t)\in\mathcal{R}$ is named diffusion coefficient, and $\bm{w}$ is the Wiener process whose derivative $d\bm{w}$ is characterized by a standard Gaussian random variable (white Gaussian noise). 
The corresponding marginal distribution $p_t(\bm{x})$ at time $t$ satisfies $p_0(\bm{x}) = p(\bm{x})$ and $p_T(\bm{x}) \approx \rho(\bm{x})$. To sample from $p(\bm{x})$, the diffusion model first samples from $\rho(\bm{x})$ and then transforms $\bm{x}$ backward according to the inverse SDE associated with  \cref{eq:prelim:forward_sde}
\begin{align} \label{eq:prelim:backward_sde}
    d\bm{x} = [f(\bm{x},t) - g^2(t)\nabla_{\bm{x}}\log p_t(\bm{x})]dt + g(t)d\bar{\bm{w}},
\end{align}
where $dt$ is the negative time flow, evolving from $t=T$ to $t=0$, and $\bar{\bm{w}}$ is the Wiener process in the inverse time direction whose derivative $d\bar{\bm{w}}$ is again a standard Gaussian random variable because of the Gaussian increment characterization of Wiener process. In \cref{eq:prelim:backward_sde}, $\nabla_{\bm{x}}\log p_t(\bm{x})$ is the unknown score function of $\bm{x}$ at time $t$, which is estimated by a neural network $s_{\bm{\theta}}(\bm{x},t)$ with the weighted score matching loss
\begin{align} \label{eq:prelim:scorematching_loss}
    &\mathcal{J}_{SM}(\bm{\theta};\lambda(\cdot)) \nonumber\\ &\quad= \frac{1}{2}\int_{0}^{T} \mathbb{E}_{p_t(\bm{x})}[\lambda(t)||\nabla_{\bm{x}}\log p_t(\bm{x}) - s_{\bm{\theta}}(\bm{x},t)||^2_2]dt.
\end{align}

\section{Unseen Distribution Transformation}\label{sec:analysis}
First, we motivate our method by analyzing why the diffusion model helps promote the OoD generalization. 

Here, we proof that, by feeding a linear combination of the OoD samples and the standard Gaussian noise to a diffusion model, the OoD samples can be transformed to the ID samples. 

\begin{theorem} \label{thm:1}
Given a diffusion model trained on the source distribution $p(x)$,let $p_t$ denote the distribution at time $t$ in the forward transformation, let $\bar{p}(x)$ denote the output distribution when the input of the backward process is standard Gaussian noise $\bm{\epsilon}$ whose distribution is denoted by $\rho(x)$, let $\omega(x)$ denote the output distribution when the input of backward process is a convex combination $(1-\alpha)X'+\alpha \bm{\epsilon}$, where random variable $X'$ is sampled following the target distribution $q(x)$ and $\alpha\in(0,1)$. Under some regularity conditions detailed in Appendix, we have
\begin{align}
    KL(p||\omega) \le \mathcal{J}_{SM} + KL(p_T||\rho) + \mathcal{F}(\alpha)
\end{align}
\end{theorem}

\cref{thm:1} proves that the testing distribution can be transformed to the source distribution and the convergence is controlled by $\alpha$.
The first terms in the inequality are introduced by the pretrained diffusion model. The loss term $\mathcal{J}_{SM}$ is small after sufficient training and can be reduced if the training procedure is further optimized. The KL-divergence term $KL(p_T||\rho)$ indicates the distance between the real distribution achieved by the forward transformation and the manually chosen standard Gaussian distribution, which is monotonically decreasing as the $T$ goes larger and converges to $0$ as the $T$ is sufficient large. The third term (detailed in Appendix) is introduced by the distribution of the OOD testing samples, which is controlled by $\alpha$ and converges to $0$ as $\alpha$ goes to $1$. 

\subsection{Implementation}

In this part, we describe \ouralg \ to realize the idea of shifting testing image towards the training distribution to boost the OoD generalization. The entire workflow is illustrated in \cref{fig:framework}.
\input{2figs/framework}

In the previous analysis, by constructing a linear combination of the OoD input and the noise, the alignment between the testing and the training distributions is converted to an alignment problem in the noise space. To be consistent with the diffusion model literature, we rewrite the linear combination as $\hat{x} = \beta x + \alpha \bm{\epsilon}$. By adjusting the coefficients $\alpha$, $\beta$ and the starting time $s$, the distance between the distribution of $\hat{x}$ (i.e., $w(\hat{x})$) and the input distribution of the diffusion model (during the training time) at time $s$ (i.e., $p_s(x)$) is controlled. To better match $w(\hat{x})$ to $p_s(x)$, utilizing the common practice~\cite{ddpm,ddim} that $x_s$ is a designed to be a linear combination of $x$ and Gaussian noise, we keep $s$ as a hyperparameter and calculate $\alpha$ and $\beta$ from $s$. Conditioned on the diffusion model, the maps from $s$ to $\alpha$ and $\beta$ vary. Taking DDPM as an example, the calculations can be written as $\alpha = \sqrt{1-\prod_{l=1}^s(1-\sigma_l)}$ and $\beta= \sqrt{\prod_{l=1}^s(1-\sigma_l)}$, where $\sigma_l$ is the standard deviation of the noise in the forward process at time $l$.

Next, we look into the schedule of starting time $s$. Two factors are taken into consideration. First is the time efficiency. The generation of the diffusion model requires an iterative sampling in which the total sampling steps (thus the sampling time) are controlled by the starting time $s$. Though providing more sufficient distribution transformation, using a large $s$ is time-consuming and inappropriate for the online testing environment.~\cite{trilemma,closerfaster} Second is the sample differences.
As discovered in image editing\cite{SDEdit} using the diffusion model, the starting time controls the faithfulness and reality of the generated images. 
Analogically, in using the diffusion model to transfer the OoD data, $s$ is a controller of the degree of distribution transformation and the preservation of semantic information, which are both crucial to OoD prediction.
In practice, the distance from different samples to the training distribution is different and the difficulty of the semantic label preservation varies among the samples, which makes a uniform starting time schedule improper.

\input{4algs/alg}

Thus, in \ouralg, a sample adaptive schedule of the starting time index is used, in which $s$ increases gradually. Specifically, given a predefined starting time sequence $\{s_l\}_{l=0}^L$, where $s_0<s_1<\cdots<s_L$, for each OoD sample $x$, \ouralg\   begins with the smallest $s_0$, uses the diffusion model to transfer the combination $\hat{x}$, and lets the predictor output the prediction result $h$ together with the confidence score $c$ of the prediction. If the $c$ is greater than a predefined threshold $k$, the prediction is accepted. 
Otherwise, \ouralg\ rejects the prediction, move to the next time index, and repeats the above procedure (transfer, prediction, and then using confidence score to make a judgment). If the prediction confidence score still does not meet the threshold at time index $s_L$, the last prediction is accepted.
By doing so, \ouralg\ allocates a small starting time index to the samples close to the source distribution and a larger one to the samples far from the source distribution. Thus, this adaptation procedure avoids the use of a uniformly small time step which  causes some samples to not be fully transferred, or a uniformly large time steps to cause the diffusion model to take too long to proceed. 
In our experiments, we use the maximum class probability as the confidence score and set the confidence threshold as hyperparameter.

\input{3tab/Main_table}

Usually, under the setting of OoD prediction, multiple training domains are available. The majority of the OoD prediction algorithms are based on this multiple-training domain assumption to extract stable features and establish a uniform predictor. Consistent with this common setting, for the multiple training domain problem, we use a unified predictor $f$ trained on all available training domains. However, a mixture of multiple distributions increases the difficulty of training for the diffusion model, which leads to a slower convergence and domain collapse (only generating samples from easy domains).
Given consideration to these undesired phenomena, for the multiple training domain problem, we train individual diffusion model $g_m$, where $m=1,\cdots,M$ is the training domain index, on each training domain, use every $g_m$ to transfer the OoD sample, and then obtain the prediction $h_m$ from every transferred sample using $f$. Finally, we ensemble the predictions $\{h_m\}_{m=1}^M$ together to get the overall prediction. In our experiments, the ensemble is conducted by averaging the logits (the inputs of the softmax layer) in the predictor neural network.

The overall framework of \ouralg\  is shown in \cref{alg}.

\input{2figs/True_PosNeg_Fig}

\section{Experiments}
\textbf{Datasets.} 
In the main text, experiments on the following datasets are reported. More experiments can be found in supplementary materials.
(1) PACS\cite{pacs} contains 999,1 colored images from 7 classes and  4 domains (art painting, cartoon, photo, and sketch); (2) Office-Home~\cite{officehome} contains around 15,500 images from 65 different classes and 4 domains (artistic, clip art, product, and real-world). 

\textbf{Base Methods.} Based on the taxonomy in \cref{sec:related_work}, we select one (or two) up-to-date or representative algorithm(s) from each category as the base methods in the experiments with multiple training domains. For experiments with a single training domain, we adjust the list of base methods by only choosing the algorithms without multi-domain assumption.  For each experimental configuration (dataset and training-testing split), the hyperparameters of each algorithm are optimized based on the performance metric on the testing set over 10 random searches. The searching regions are reported in supplementary materials. Fixing the optimal hyperparameters, we repeat the experimental pipeline three times and report the average results together with the standard deviations to alleviate the influence of the lucky weight initialization and dataset split. 

 In \cref{tab:main,tab:pacs_single}, we use [algorithm]* to indicate the use of \ouralg, the \mycolorbox{green cells} to indicate our method improves the base method, and the \mycolorbox{\textbf{green cells with text in bold}} to indicate our method improves the base method with non-overlapped confidence interval.

\subsection{Multi-training Domain Generalization}
The experiments for multi-training domain generalization are conducted on PACS and OfficeHome. For each dataset, one domain is left as the testing domain and the other three are used for training. The performance comparison of base methods with and without our method are listed in \cref{tab:main}. Our method generally enhances the OoD prediction of all types of base methods by 3.68\% on PACS and 0.84\% on OfficeHome, which indicates \ouralg\ successfully closes the distance between the OoD samples and the training distribution. 
The significance of the improvements varies among different leave-one-out settings, because the accuracy of the base methods varies. When the accuracy of the base method is low (e.g., the average accuracy of the base methods is 61.88\% for S of PACS), more (8.22\%) improvement is achieved. When the accuracy of the base method is high (e.g., the average accuracy of the base methods is 94.61\% for P of PACS), the absolute value of the improvement is less (0.89\%). 
We also notice that the rise in performance on OfficeHome dataset is less than the PACS, which can be attributed to that the larger number of classes in OfficeHome increases the hardness of the semantic label preservation of our method. 

We further analyze whether the performance gain depends on a certain class or domain. We count, for each class in each testing domain, (1) the number of samples correctly predicted by both the base methods and the base methods with \ouralg ($\#\ Both\ Correct$), (2) the number of samples correctly predicted by the base methods with \ouralg\  but wrongly predicted by base methods ($\#\ Only\ Ours\ Correct$), (3) the number of samples correctly predicted by the base methods ($\#\ Base\ Correct$), and (4) the number of samples wrongly predicted by the base methods ($\#\ Base\ Wrong$). \cref{fig:true_pn} shows the preservation ratio calculated by $\frac{\#\ Both\ Correct}{\#\ Base\ Correct}$ and the correction ratio calculated by $\frac{\#\ Only\ Ours\ Correct}{\#\ Base\ Wrong}$. The preservation ratio indicates the percentage of the correct predictions that are still correct when our method is used.  The correction ratio indicates the percentage of the wrong predictions that are corrected when our method is used. Averaging over testing domains and classes, 94.52\% of the correct predictions are still correct when our method is used. The highest preservation ratio is 95.95\% appearing when photo is the testing domain. Among the testing domains and the classes, our method can at least correct 8.57\% wrong predictions and averagely 28.17\%. These results prove the general effectiveness of our methods across different domains and classes.

\input{3tab/PACS_single}
\subsection{Single-training Domain Generalization}
To cover the harder single training domain generalization, we do experiments on PACS datasets in which we use each domain as the training domain and test on the other three.
The performance comparison of base methods with and without our method on the dataset with single training domains are listed in \cref{tab:pacs_single}. With our method, irrelevant to the training and testing domain, general performance gains are achieved and the average performance gain is 2.42\%.

\input{2figs/proper_s}

\subsection{Discussion on the starting time $s$}

Transforming the distribution closer to the source distribution increases the confidence of the prediction based on the transformed image. On the other hand, the well-preserved label information guarantees the prediction of the generated image has the same label as its origin. However, increasing $s$ to get further distribution transformation will destroy the label information, contrarily, decreasing $s$ to preserve the label information will limit the distribution transformation. Thus, an optimal $s$ exists for each algorithm. (see \cref{fig:proper_s})

\input{2figs/hyper}
\subsection{Discussions on hyperparameters}
\textbf{Ensemble.} We analyze whether only transforming the testing domain data to a subset of the training domains influences the performance gain or not. \cref{fig:hyper:a} shows the accuracy with different ensemble sets on PACS with the multi-training domain setting and Cartoon is the testing domain. Each row corresponds to one kind of ensemble. For example, row ``C'' shows the accuracy when the prediction is only based on the original testing image, which is the accuracy of the base methods; row ``C,A'' shows the accuracy when the prediction based on the original testing image and prediction based on the image transformed towards the domain Art are considered together. As shown from the figure, including more components in the ensemble set increases the average performance and reduces its variance; ensembles without the original prediction can perform comparably with the base method; the ensembles with the original prediction and any (one, two, or all) of the predictions based on the transformed image is statistically better than the base methods. 

\textbf{Confidence Threshold.} We analyze the influence of the confidence threshold $k$, and the performance with varying $k$ are shown in \ref{fig:hyper:b}. Base methods are special cases when $k=0$, their performances are shown along the \textcolor{myred}{red vertical line}
Though tuning $k$ enlarges the performance gain, when $k>0$, our method is activated and there are always performance improvements.

\vspace{-0.5em}
\section{Conclusion}
\vspace{-0.5em}

In this paper, we investigate a novel task, termed unseen distribution transformation, which aims at transforming the unseen distribution towards the seen distribution. 
We frame the first solution to unseen distribution transformation, in which the unseen distribution is first linearly combined with the Gaussian noise and then transformed by a diffusion model trained on the seen domain. 
By solving this task, we provide a new perspective for addressing the OoD prediction task by first closing the distance between the testing domain and the training domain and then drawing a prediction. 
We propose the portable \ouralg, which conducts sample adaptive unseen distribution transformation to enhance the OoD prediction algorithms.
Experimental results show that our method results in general performance gains when inserted into various types of base methods under multi-training and single-training domain generalization problems.

\section{Acknowledgment}
This project is supported by the Singapore Ministry of Education Academic Research Fund Tier 1~(Grantor’s Reference Number: 23-0306-A0001), a project titled ``Towards Robust Single Domain Generalization in Deep Learning''.

%% file: 2figs/toy_example.tex
\begin{figure}
  \centering
    \includegraphics[width=\linewidth]{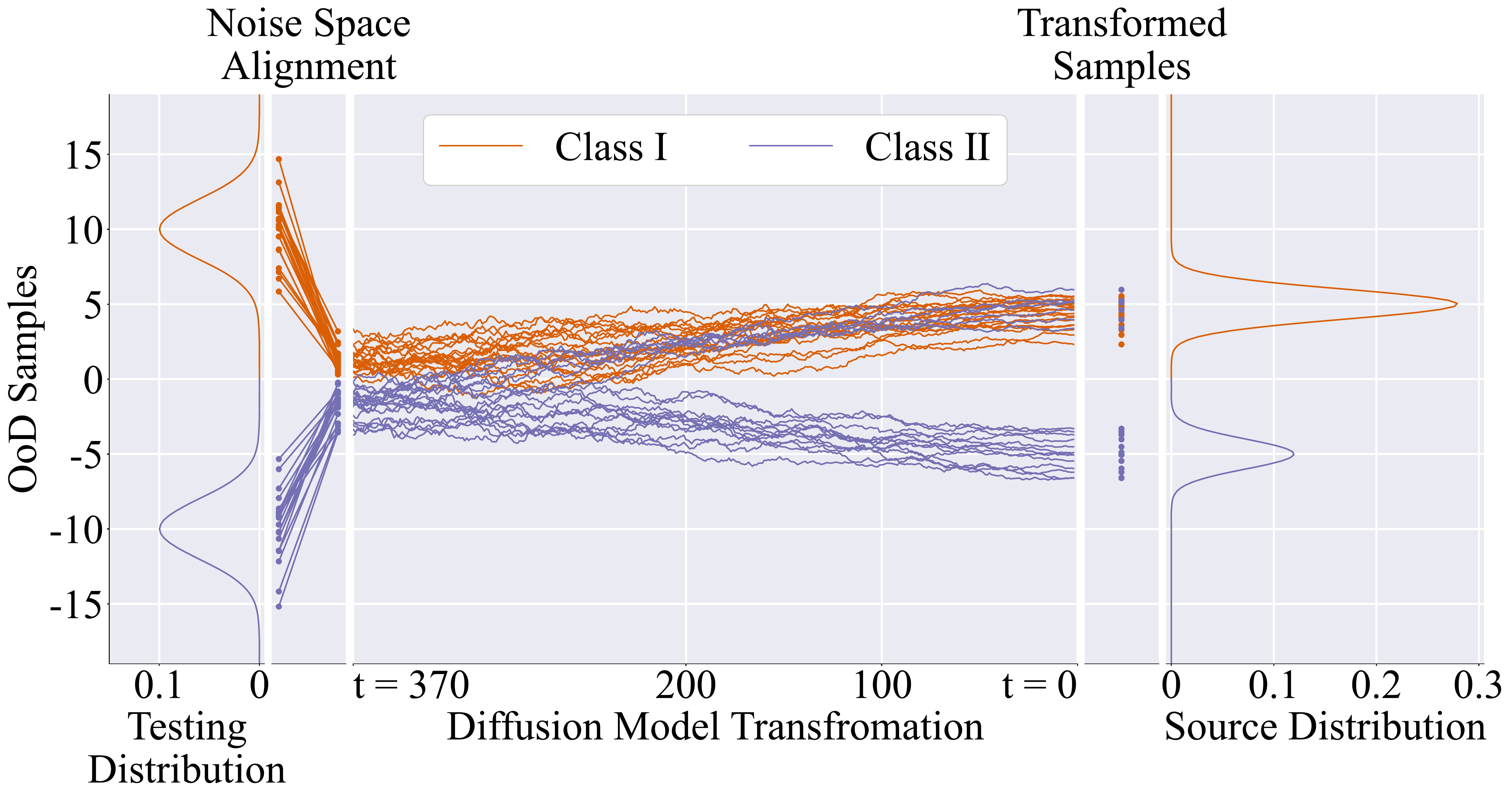}
    \caption{Using a Diffusion Model to solve the one-dimensional UDT. OoD samples are transformed to the source distribution with limited failure of label preservation. }
  \label{fig:toy}
  \vspace{-2em}
\end{figure}

%% file: 2figs/framework.tex
\begin{figure*}
  \centering
    \includegraphics[width=\linewidth]{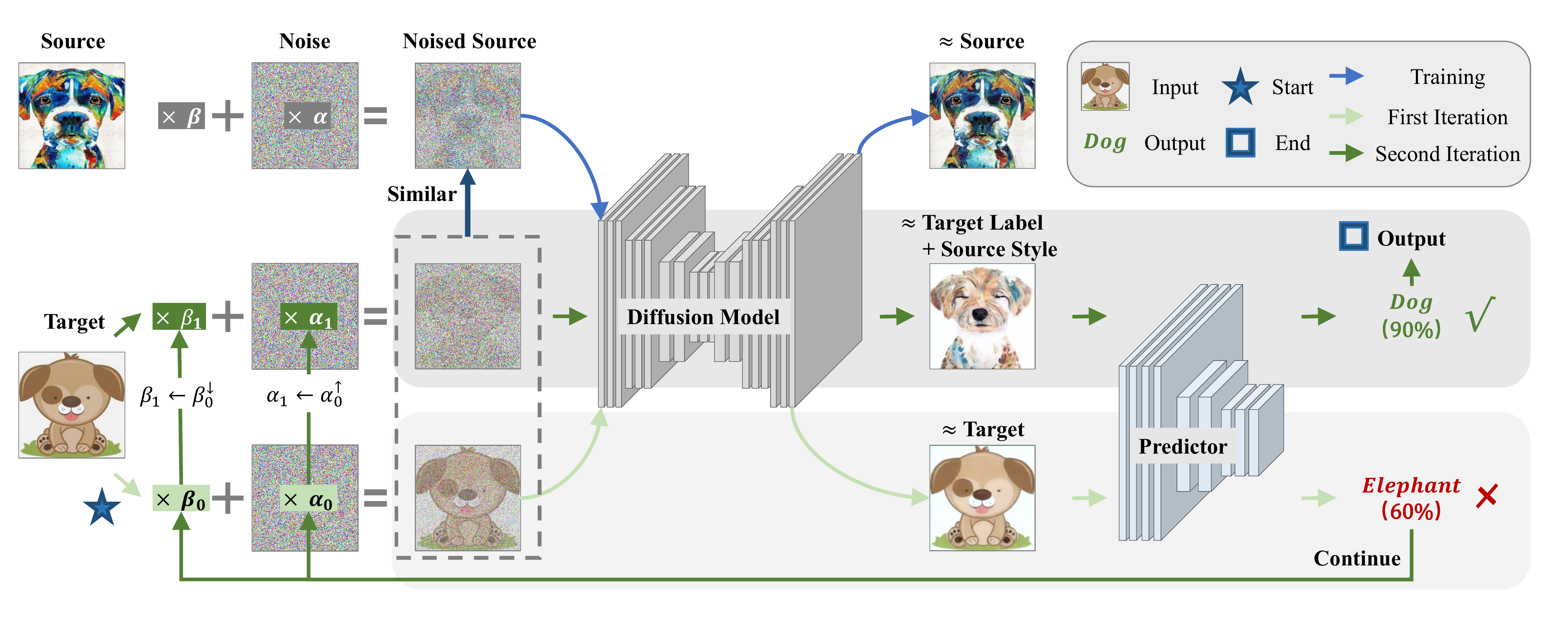}
  \caption{Illustrative example of DSI finished in two iterations. In the first iteration, insufficient transformation leads to wrong prediction with low confidence. Our algorithm rejects the prediction and continues the second iteration. In the second iteration, with proper transformation, the correct prediction is drawn and accepted. Then, the algorithm finished.}
  \label{fig:framework}
\end{figure*}

%% file: 4algs/alg.tex
\begin{algorithm}
\caption{Distribution Shift Inversion}\label{alg}
\begin{algorithmic}[1]
\Require Predictor: $f$, Diffusion Models: $\{g_m\}_{m=1}^M$, Function for calculating $\alpha$ and $\beta$: $\phi$, Ensemble function: $\xi$, Confidence score function: $\psi$
\Require Starting time series: $\{s_l\}_{l=0}^L$, Threshold: $k$
\Require OoD sample: $x$
\State $h_0 \gets f(x)$ \Comment{Draw prediction from original sample}
\For{$l$ \textbf{in} $1,\cdots,L$} 
\State $\alpha,\beta \gets \phi(t_l)$
\State $\hat{x} \gets \beta x+ \alpha \epsilon$
\For{$m$ \textbf{in} $1,\cdots,M$} 
\State $\tilde{x} \gets g_m(\hat{x})$ \Comment{Transfer}
\State $h_m \gets f(\tilde{x})$ 
\EndFor
\State $h \gets \xi(h_0,\cdots,h_M)$  \Comment{Ensemble}
\If{$\psi(h)>k$} \Comment{Confidence Filtering}
    \State \textbf{break}
\EndIf
\EndFor
\State \textbf{output prediction} $h$ \Comment{Accept the last prediction}
\end{algorithmic}
\end{algorithm}

%% file: 3tab/Main_table.tex
\begin{table*}
    \scriptsize
    \centering
\begin{NiceTabular}{@{}l@{$\quad\ \ $}r@{$\pm$}l@{$\quad\ \ $}r@{$\pm$}l@{$\quad\ \ $}r@{$\pm$}l@{$\quad\ \ $}r@{$\pm$}l@{$\quad\ \ $}r@{$\pm$}l@{$\quad\ \ $}r@{$\pm$}l@{$\quad\ \ $}r@{$\pm$}l@{$\quad\ \ $}r@{$\pm$}l@{$\quad\ \ $}r@{$\pm$}l@{$\quad\ \ $}r@{$\pm$}l@{}}[colortbl-like]
\toprule
Dataset     & \multicolumn{10}{c}{PACS}                                                                                                     & \multicolumn{10}{c}{OfficeHome}                                                                                             \\ \midrule
Testing Domain & \multicolumn{2}{c}{A} & \multicolumn{2}{c}{C} & \multicolumn{2}{c}{P} & \multicolumn{2}{c}{S} & \multicolumn{2}{c}{Average}   & \multicolumn{2}{c}{A} & \multicolumn{2}{c}{C} & \multicolumn{2}{c}{P} & \multicolumn{2}{c}{R} & \multicolumn{2}{c}{Average} \\ \midrule
ERM                      &76.24		&0.44		&67.90		&0.64		&94.47		&0.44		&60.32		&0.85		&74.73		&0.59		&60.81		&0.51		&54.17		&0.17		&74.80		&0.35		&76.11		&0.40		&66.47		&0.36\\
ERM*                     &\u{78.52}	&\u{0.81}	&\u{70.90}	&\u{0.76}	&\u{95.80}	&\u{0.16}	&\u{67.97}	&\u{0.68}	&\u{78.30}	&\u{0.60}	&\g61.15	&\g0.56	    &\u{55.96}  &\u{0.76}	&\u{75.55}	&\u{0.23}	&\g76.56	&\g0.29	    &\u{67.31}	&\u{0.46}\\
ANDMask~\cite{andmask}   &67.35		&0.51		&63.22		&2.23		&93.98		&0.26		&60.35		&1.05		&71.23		&1.01		&60.84		&0.60		&55.73		&0.60		&73.80		&0.44		&75.88		&0.08		&66.56		&0.43\\
ANDMask*                 &\u{69.43}	&\u{0.40}	&\u{68.23}	&\u{2.29}	&\u{94.99}	&\u{0.24}	&\u{69.63}	&\u{0.57}	&\u{75.57}	&\u{0.88}	&\g61.18	&\g0.65	    &\u{56.80}	&\u{0.41}	&\g74.35	&\g0.40	    &\u{76.40}	&\u{0.54}	&\g67.18	&\g0.50\\
CAD~\cite{cad}           &75.42		&0.51		&68.20		&0.64		&95.02		&0.55		&59.77		&2.14		&74.60		&0.96		&60.71		&0.64		&53.39		&0.18		&74.41		&0.29		&75.36		&0.39		&65.97		&0.38\\
CAD*                     &\g76.11	&\g0.81		&\u{71.81}	&\u{0.59}	&\u{96.09}	&\u{0.16}	&\u{69.56}	&\u{1.69}	&\u{78.39}	&\u{0.81}	&\g61.04	&\g0.63	    &\u{55.08}	&\u{0.44}	&\u{75.20}	&\u{0.29}	&\u{76.14}	&\u{0.28}	&\u{66.87}	&\u{0.41}\\
CondCAD~\cite{cad}       &75.55		&0.38		&67.77		&0.50		&95.57		&0.12		&59.99		&1.90		&74.72		&0.73		&60.74		&0.37		&55.27		&0.83		&74.64		&0.26		&75.42		&0.20		&66.52		&0.42\\
CondCAD*                 &\u{77.12}	&\u{0.36}	&\u{71.61}	&\u{0.37}	&\u{96.03}	&\u{0.09}	&\u{69.79}	&\u{1.96}	&\u{78.64}	&\u{0.70}	&\g61.15	&\g0.28		&\g56.09	&\g0.58	    &\u{75.52}	&\u{0.41}	&\u{76.43}	&\u{0.17}	&\u{67.30}	&\u{0.36}\\
GroupDRO~\cite{groupDRO} &69.56		&1.01		&63.77		&3.74		&93.16		&0.77		&64.16		&1.78		&72.66		&1.83		&60.32		&0.26		&53.12		&0.64		&71.61		&0.66		&75.68		&0.21		&65.18		&0.44\\
GroupDRO*                &\u{71.61}	&\u{0.81}	&\g67.45	&\g2.75		&\g93.78	&\g0.40		&\u{71.88}	&\u{1.02}	&\u{76.18}	&\u{1.25}	&\g60.73	&\g0.26		&\u{55.83}	&\u{0.79}	&\u{73.01}	&\u{0.92}	&\u{76.14}	&\u{0.18}	&\u{66.43}	&\u{0.54}\\
IB\_ERM~\cite{ibirm}     &77.18		&0.20		&69.76		&0.24		&94.82		&0.80		&61.26		&0.58		&75.76		&0.46		&58.30		&0.78		&54.39		&0.10		&70.36		&0.05		&72.02		&0.44		&63.77		&0.34\\
IB\_ERM*                 &\u{79.07}	&\u{0.20}	&\u{71.88}	&\u{0.44}	&\g95.83	&\g0.30		&\u{69.95}	&\u{0.88}	&\u{79.18}	&\u{0.46}	&\g58.66	&\g0.81		&\u{55.66}	&\u{0.39}	&\u{71.29}	&\u{0.01}	&\u{72.90}	&\u{0.34}	&\u{64.63}	&\u{0.39}\\
IB\_IRM~\cite{ibirm}     &75.26		&1.09		&67.77		&0.97		&94.76		&0.45		&60.25		&2.40		&74.51		&1.23		&58.01		&0.44		&52.60		&1.00		&72.75		&0.60		&74.90		&0.65		&64.57		&0.67\\
IB\_IRM*                 &\g77.31	&\g1.08		&\u{70.70}	&\u{0.63}	&\u{95.96}	&\u{0.12}	&\u{68.03}	&\u{1.67}	&\u{78.00}	&\u{0.88}	&\g58.37	&\g0.39		&\g53.68	&\g0.44		&\g72.95	&\g0.84		&\g75.55	&\g0.79		&\g65.14	&\g0.62\\
Mixup~\cite{mixup}       &71.29		&0.49		&66.11		&2.32		&94.08		&0.47		&64.81		&2.88		&74.07		&1.54		&61.46		&0.20		&55.83		&0.65		&74.06		&0.44		&76.73		&0.41		&67.02		&0.43\\
Mixup*                   &\u{73.18}	&\u{0.17}	&\g69.63	&\g2.35		&\g94.56	&\g0.40		&\u{71.45}	&\u{2.74}	&\u{77.21}	&\u{1.42}	&\u{62.25}	&\u{0.31}	&\u{57.85}	&\u{0.85}	&\u{75.10}	&\u{0.40}	&\g77.21	&\g0.26		&\u{68.10}	&\u{0.46}\\
SANDMask~\cite{sandmask} &67.35		&0.51		&62.50		&3.85		&94.17		&0.62		&64.58		&0.58		&72.15		&1.39		&61.78		&0.36		&55.24		&0.12		&73.34		&0.40		&75.68		&0.37		&66.51		&0.31\\
SANDMask*                &\u{69.43}	&\u{0.40}	&\g67.42	&\g2.96		&\g94.73	&\g0.42		&\u{72.66}	&\u{0.63}	&\u{76.06}	&\u{1.10}	&\g62.14	&\g0.35		&\u{56.58}	&\u{0.28}	&\g73.50	&\g0.23		&\g76.27	&\g0.29		&\u{67.12}	&\u{0.29}\\
SelfReg~\cite{selfreg}   &76.86		&0.08		&68.62		&0.74		&96.09		&0.14		&61.39		&0.81		&75.74		&0.44		&62.50		&0.76		&56.09		&0.23		&75.52		&0.38		&75.75		&0.24		&67.47		&0.40\\
SelfReg*                 &\u{78.61}	&\u{0.41}	&\u{71.55}	&\u{0.74}	&\g96.39	&\g0.21		&\u{71.19}	&\u{0.21}	&\u{79.44}	&\u{0.39}	&\g62.90	&\g0.82		&\u{57.52}	&\u{0.14}	&\u{76.37}	&\u{0.37}	&\u{76.79}	&\u{0.20}	&\u{68.40}	&\u{0.38}\\\midrule
Average Gain             &     1.83 &    0.43 &    3.55 &    0.85&    0.80 &    0.34  &     8.52 &     1.06  &     3.68 &   0.67              &      0.41&      0.13 &    1.52&      0.52  &    0.75&      0.35  &     0.69&      0.21  &   0.84 &   0.30                  \\ 
\bottomrule
\end{NiceTabular}
  \caption{The average accuracy $\pm$ the standard deviation of base algorithms w/o our method on PACS and OfficeHome datasets. The performance is generally boosted when our method is plugged in, whichever base algorithm is  used.}
  \label{tab:main}
  \vspace{-1em}
\end{table*}

%% file: 2figs/True_PosNeg_Fig.tex
\begin{figure*}
  \centering
  \begin{subfigure}{0.24\linewidth}
    \includegraphics[width=\linewidth]{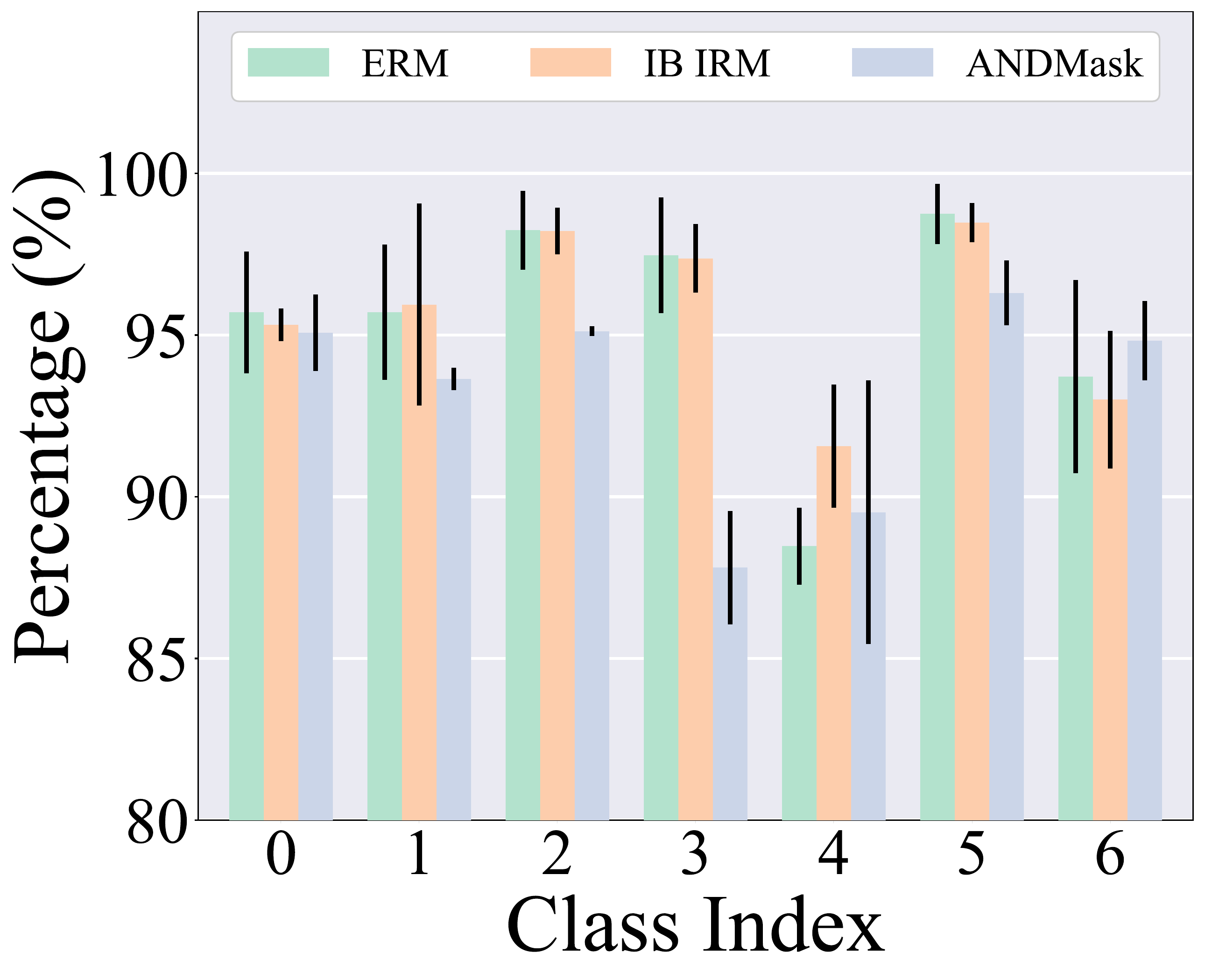}
    \caption{Testing Domain: Art}
    \label{fig:true_pn:a}
  \end{subfigure}
  \begin{subfigure}{0.24\linewidth}
    \includegraphics[width=\linewidth]{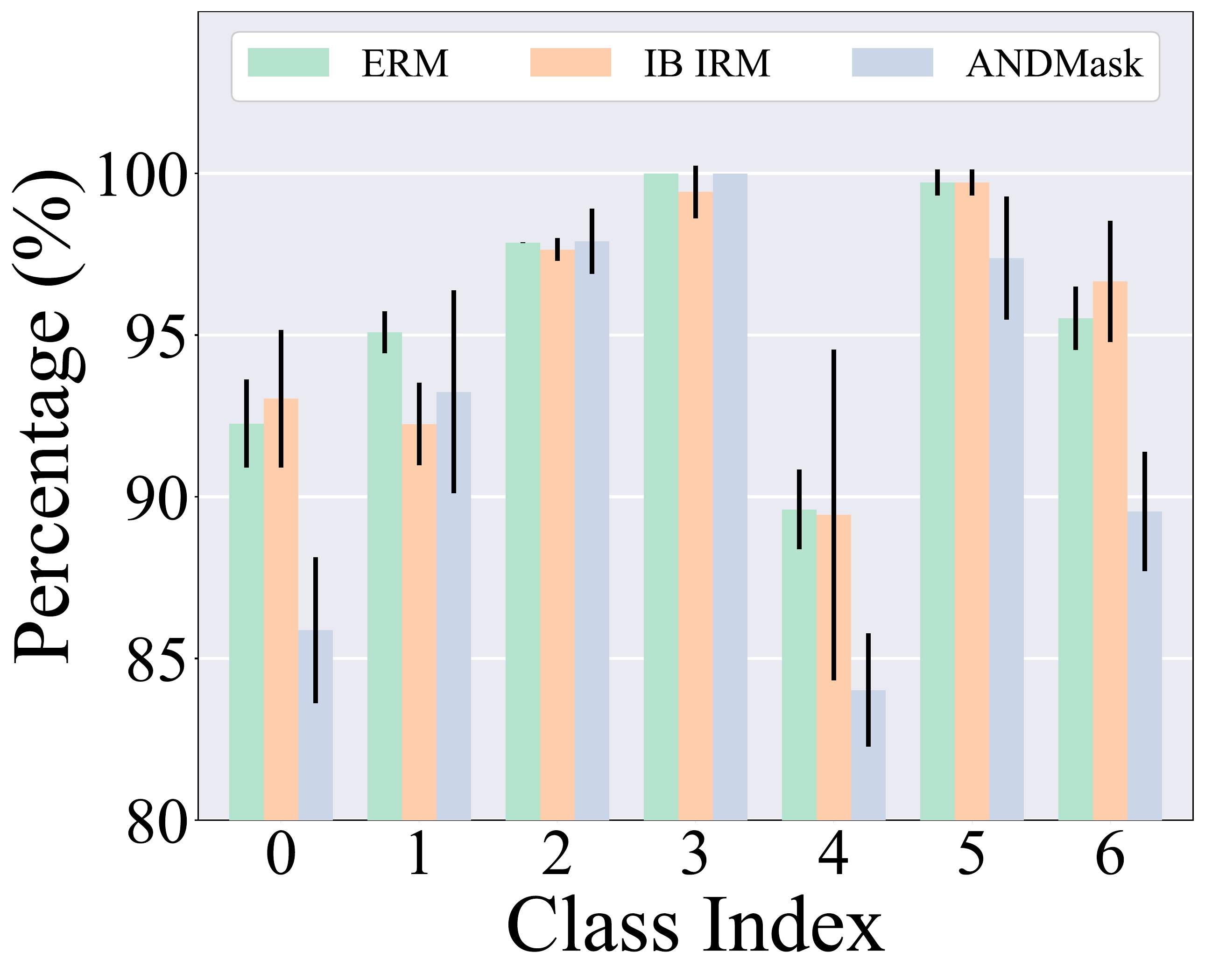}
    \caption{Testing Domain: Cartoon}
    \label{fig:true_pn:b}
  \end{subfigure}
  \begin{subfigure}{0.24\linewidth}
    \includegraphics[width=\linewidth]{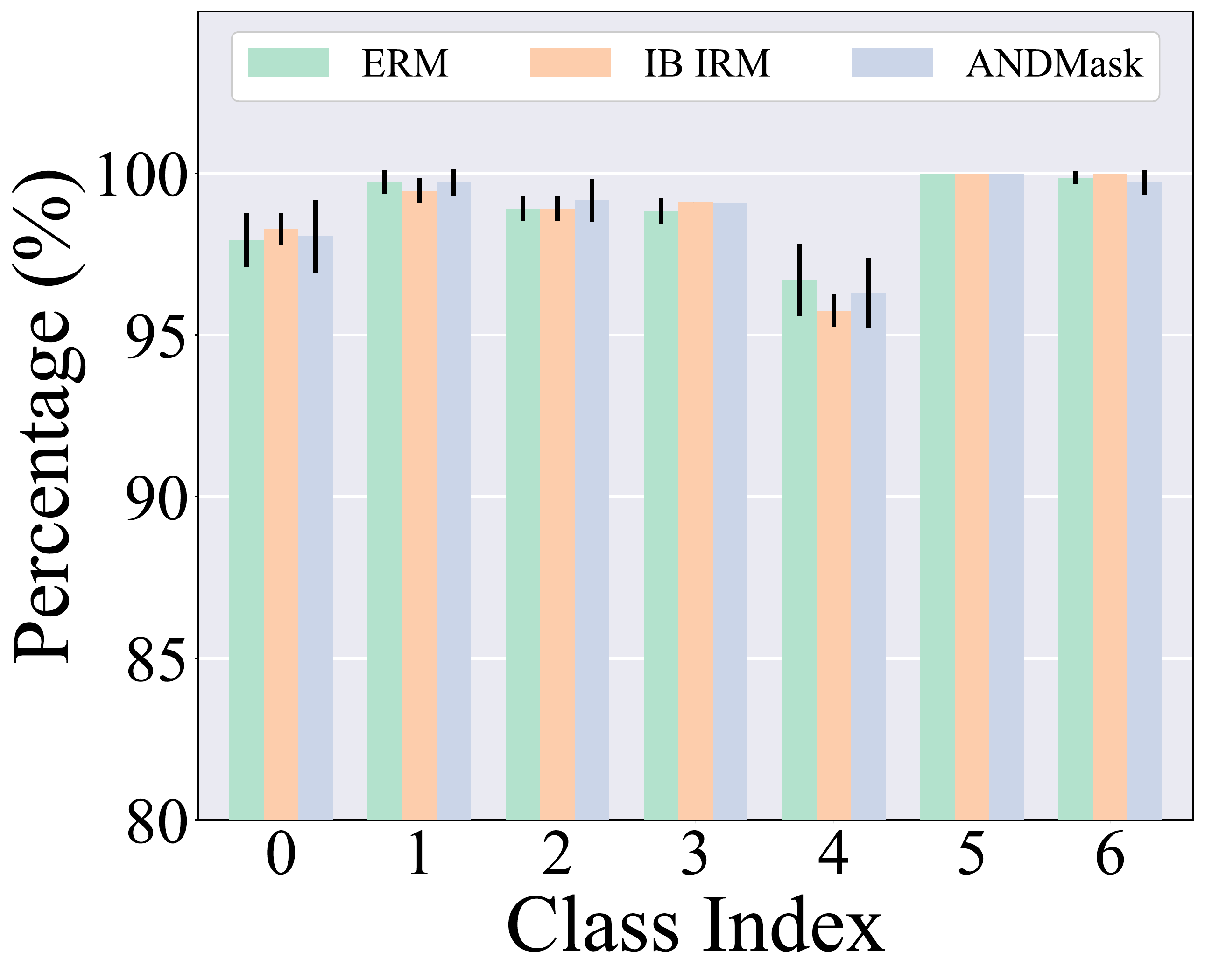}
    \caption{Testing Domain: Photo}
    \label{fig:true_pn:c}
  \end{subfigure}
  \begin{subfigure}{0.24\linewidth}
    \includegraphics[width=\linewidth]{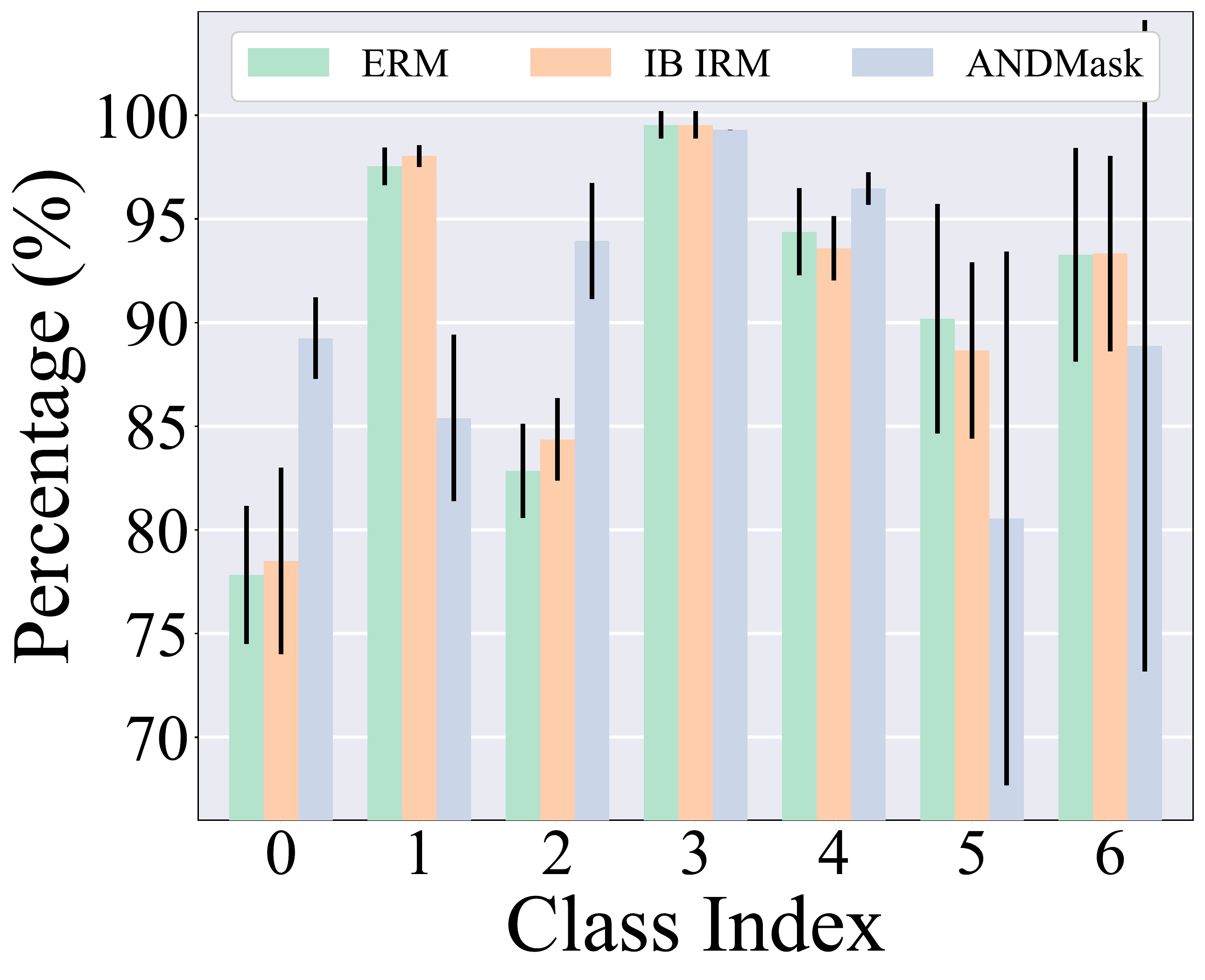}
    \caption{Testing Domain: Sketch}
    \label{fig:true_pn:d}
  \end{subfigure}
  \vspace{1em}
  \begin{subfigure}{0.24\linewidth}
    \includegraphics[width=\linewidth]{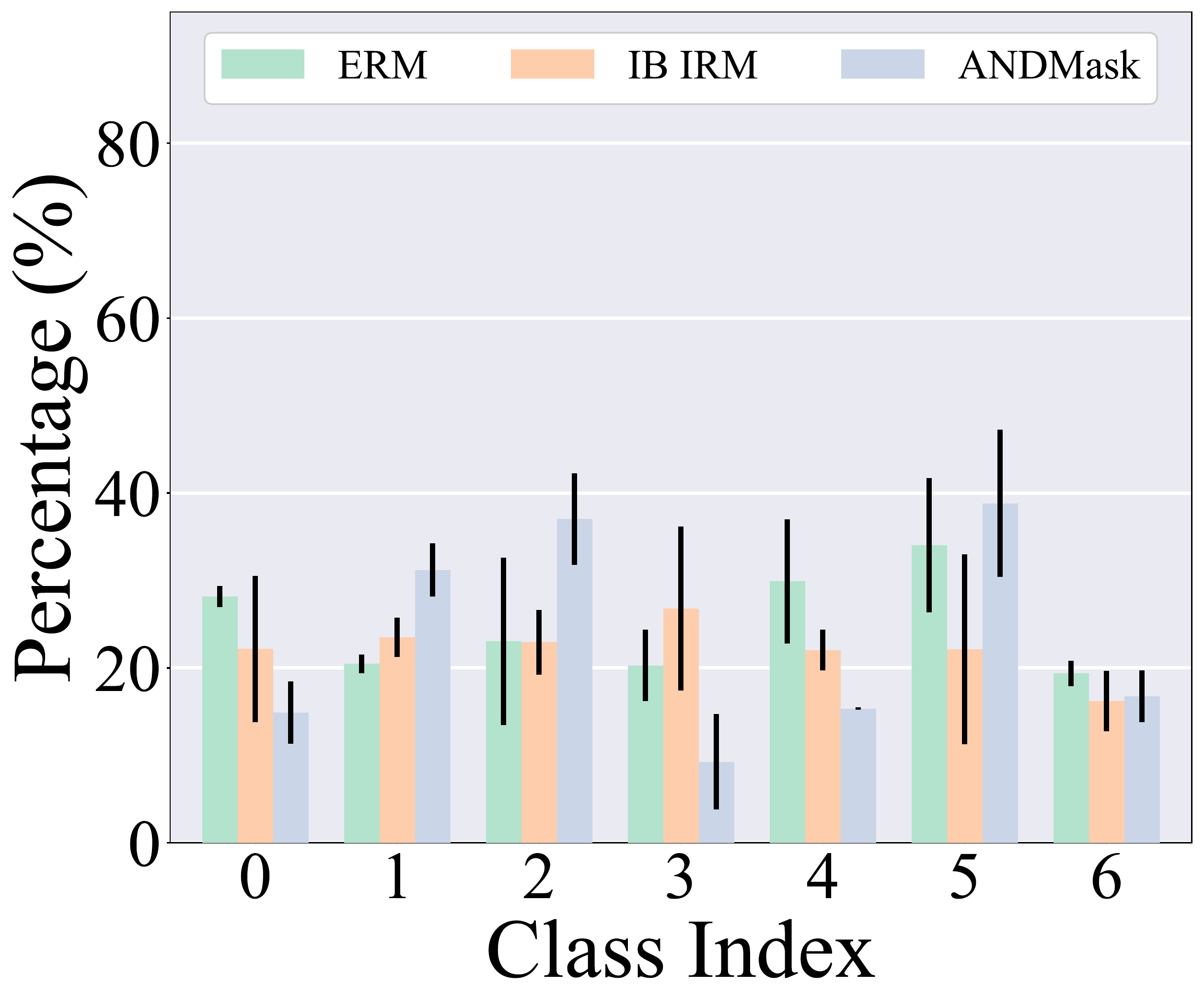}
    \caption{Testing Domain: Art}
    \label{fig:false_pn:a}
  \end{subfigure}
  \begin{subfigure}{0.24\linewidth}
    \includegraphics[width=\linewidth]{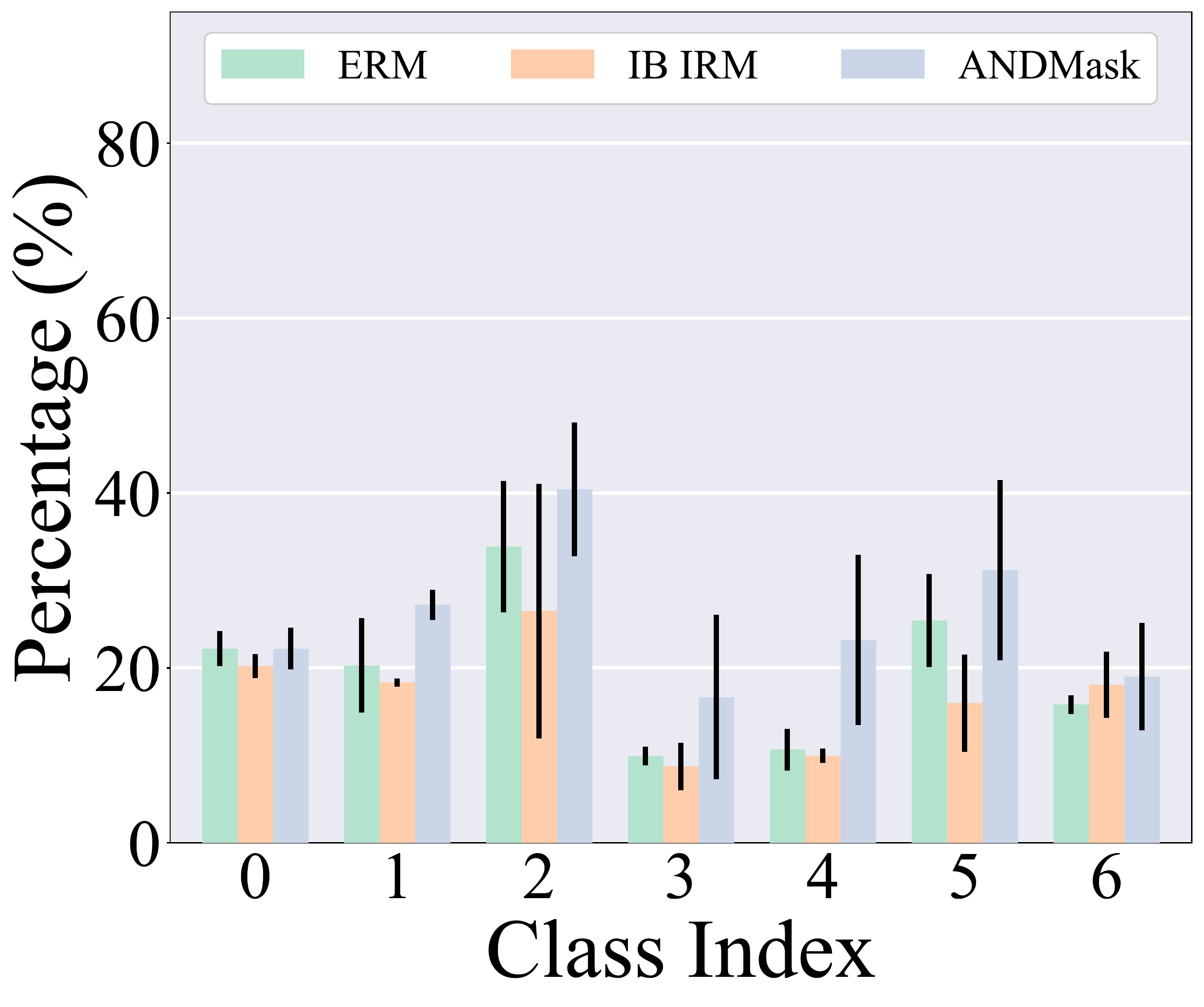}
    \caption{Testing Domain: Cartoon}
    \label{fig:false_pn:b}
  \end{subfigure}
  \begin{subfigure}{0.24\linewidth}
    \includegraphics[width=\linewidth]{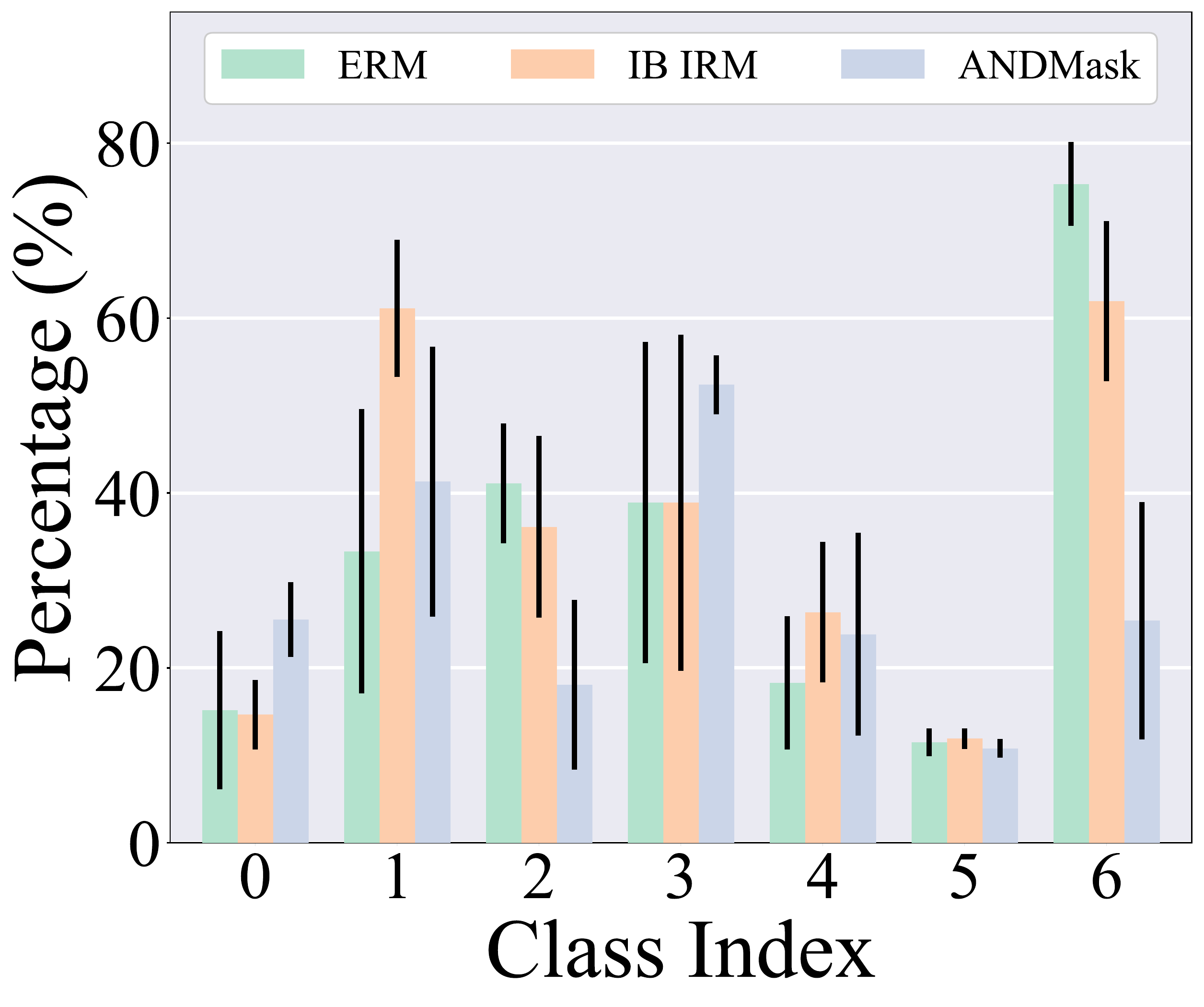}
    \caption{Testing Domain: Photo}
    \label{fig:false_pn:c}
  \end{subfigure}
  \begin{subfigure}{0.24\linewidth}
    \includegraphics[width=\linewidth]{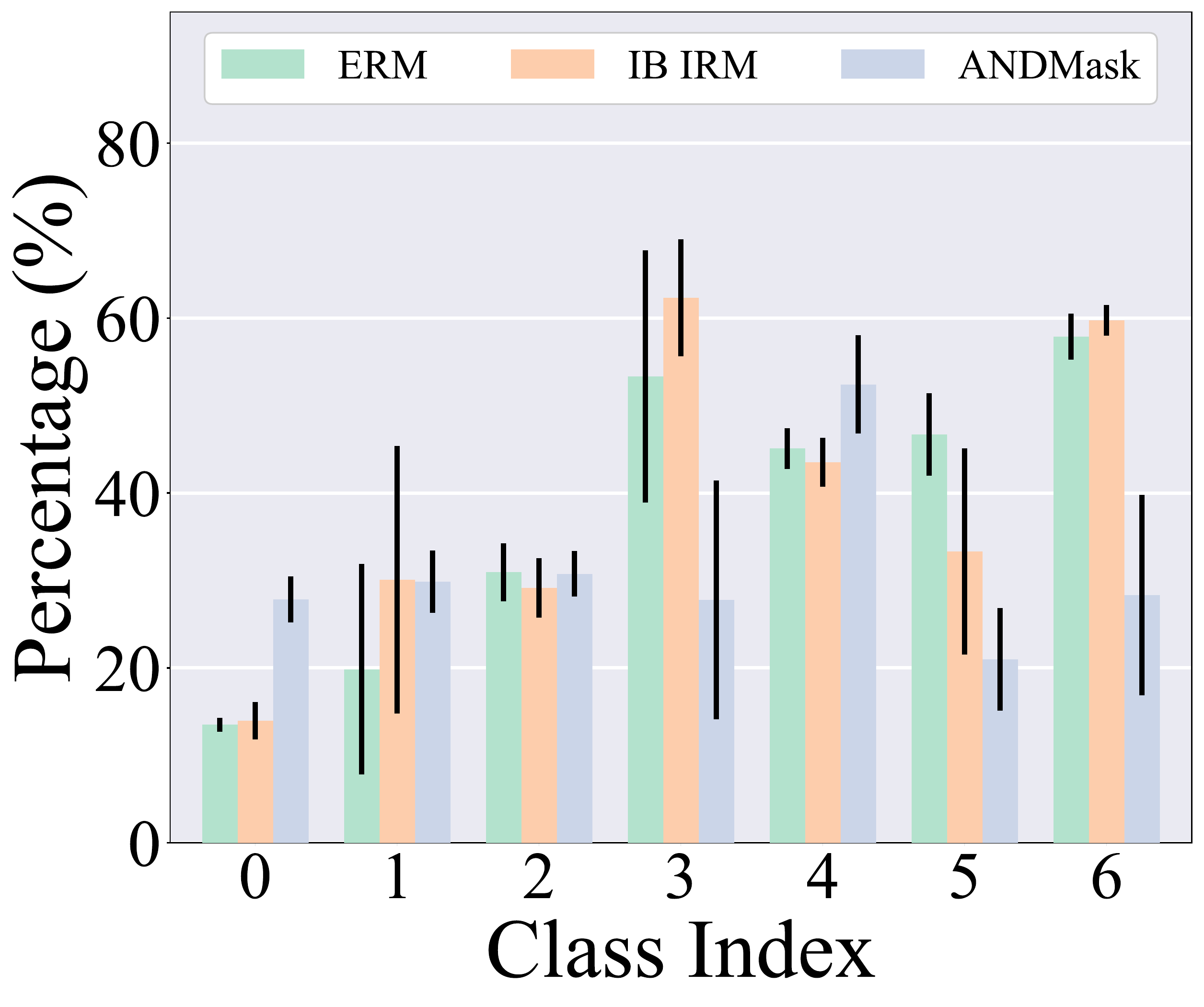}
    \caption{Testing Domain: Sketch}
    \label{fig:false_pn:d}
  \end{subfigure}
  \vspace{-1.5em}
  \caption{The percentage of correct prediction when using our method among the samples which are correctly predicted by the base method (in \cref{fig:true_pn:a} to \cref{fig:true_pn:d}), or wrongly predicted by the base method (in \cref{fig:false_pn:a} to \cref{fig:false_pn:d}). The plot is separately drawn for each leave-one-out setting and each class.}
  \label{fig:true_pn}
  \vspace{-1.5em}
\end{figure*}

%% file: 3tab/PACS_single.tex
\begin{table*}
    \scriptsize
    \centering
\begin{NiceTabular}{@{}l@{$\ \ $}r@{$\pm$}l@{$\ \ $}r@{$\pm$}l@{$\ \ $}r@{$\pm$}l@{$|$}r@{$\pm$}l@{$\ \ $}r@{$\pm$}l@{$\ \ $}r@{$\pm$}l@{$|$}r@{$\pm$}l@{$\ \ $}r@{$\pm$}l@{$\ \ $}r@{$\pm$}l@{$|$}r@{$\pm$}l@{$\ \ $}r@{$\pm$}l@{$\ \ $}r@{$\pm$}l@{}}[colortbl-like]
\toprule
Dataset         & \multicolumn{24}{c}{PACS}                                                                                                                                                                                                                                                                     \\ \midrule
Training Domain & \multicolumn{6}{c}{A}                                                 & \multicolumn{6}{c}{C}                                                 & \multicolumn{6}{c}{P}                                                 & \multicolumn{6}{c}{S}                                                 \\ \midrule
Testing Domain  & \multicolumn{2}{c}{C} & \multicolumn{2}{c}{P} & \multicolumn{2}{c}{S} & \multicolumn{2}{c}{A} & \multicolumn{2}{c}{P} & \multicolumn{2}{c}{S} & \multicolumn{2}{c}{A} & \multicolumn{2}{c}{C} & \multicolumn{2}{c}{S} & \multicolumn{2}{c}{A} & \multicolumn{2}{c}{C} & \multicolumn{2}{c}{P} \\ \midrule
ERM                 &     55.37&      0.32&     97.92&      0.05&     42.84&      1.68&     55.63&      0.26&     83.43&      0.32&     57.42&      2.27&     64.84&      1.02&     28.26&      3.86&     23.24&      4.88&     15.82&      1.33&     14.26&      0.70&     12.66&      0.49\\
ERM*                & \u{57.58}&  \u{0.26}&     97.90&      0.03& \u{48.54}&  \u{1.12}& \u{57.03}&  \u{0.44}&   \g83.98&    \g0.56& \u{63.74}&  \u{2.00}&   \g65.92&    \g0.37&   \g32.75&    \g3.78&   \g25.68&    \g4.78&   \g16.41&    \g1.51& \u{16.80}&  \u{0.76}&   \g12.92&    \g0.20\\
CORAL               &     56.35&      2.63&     96.68&      0.63&     42.32&      2.60&     55.47&      0.52&     84.15&      0.70&     57.81&      1.65&     62.04&      0.74&     32.16&      5.84&     26.33&      8.46&     15.27&      0.68&     15.76&      1.16&     11.65&      0.81\\
CORAL*              &   \g58.76&    \g2.96&   \g96.84&    \g0.53& \u{47.56}&  \u{1.16}& \u{56.87}&  \u{0.09}&   \g84.38&    \g0.83& \u{64.65}&  \u{2.01}& \u{63.90}&  \u{1.01}&   \g35.97&    \g4.80&   \g29.23&    \g7.99&   \g15.98&    \g0.66&   \g16.50&    \g1.04&   \g12.27&    \g0.92\\
RSC                 &     55.21&      0.28&     98.18&      0.17&     44.66&      1.69&     54.36&      1.06&     83.79&      0.62&     59.41&      2.33&     63.05&      0.96&     31.61&      6.75&     24.90&      5.12&     16.57&      0.59&     16.37&      0.26&     13.77&      1.92\\
RSC*                & \u{57.39}&  \u{0.45}&   \g98.24&    \g0.15& \u{50.39}&  \u{1.25}&   \g55.70&    \g0.74&   \g83.95&    \g0.56& \u{65.98}&  \u{1.60}&   \g64.13&    \g1.64&   \g35.29&    \g5.77&   \g28.22&    \g4.87&   \g17.68&    \g0.68& \u{17.45}&  \u{0.69}&   \g14.49&    \g1.88\\
SagNet              &     56.41&      0.47&     97.95&      0.08&     46.81&      3.33&     58.82&      1.18&     86.95&      0.79&     57.91&      2.62&     67.32&      0.99&     41.76&      2.35&     35.77&      2.48&     15.76&      0.91&     14.23&      2.45&     16.76&      1.51\\
SagNet*             & \u{58.85}&  \u{0.51}&   \g97.96&    \g0.08& \u{53.48}&  \u{3.30}&   \g59.51&    \g1.40&   \g87.14&    \g0.97& \u{64.71}&  \u{3.50}&   \g67.74&    \g0.75&   \g44.34&    \g2.00&   \g38.64&    \g1.84& \u{18.03}&  \u{0.67}&   \g15.95&    \g2.17&   \g18.26&    \g1.90\\
\midrule
Average Gain    &      2.31&      0.12   &     0.05&      0.07     &   5.83&      0.52   &    1.20&      0.30    &     0.28&      0.16   &     6.63&      0.21  &     1.11&      0.51  &     3.64&      0.69   &    2.88&      0.31  &     1.17&      0.67    &     1.52&      0.68    &    0.77&      0.45    \\ \bottomrule
\end{NiceTabular}
  \caption{The average accuracy $\pm$ the standard deviation of base algorithms w/o our method on PACS datasets with single training domain setting. The performance is generally boosted when our method is plugged in, whichever base algorithm used.}
  \label{tab:pacs_single}
  \vspace{-2em}
\end{table*}

%% file: 2figs/proper_s.tex
\begin{figure}
  \centering
    \includegraphics[width=\linewidth]{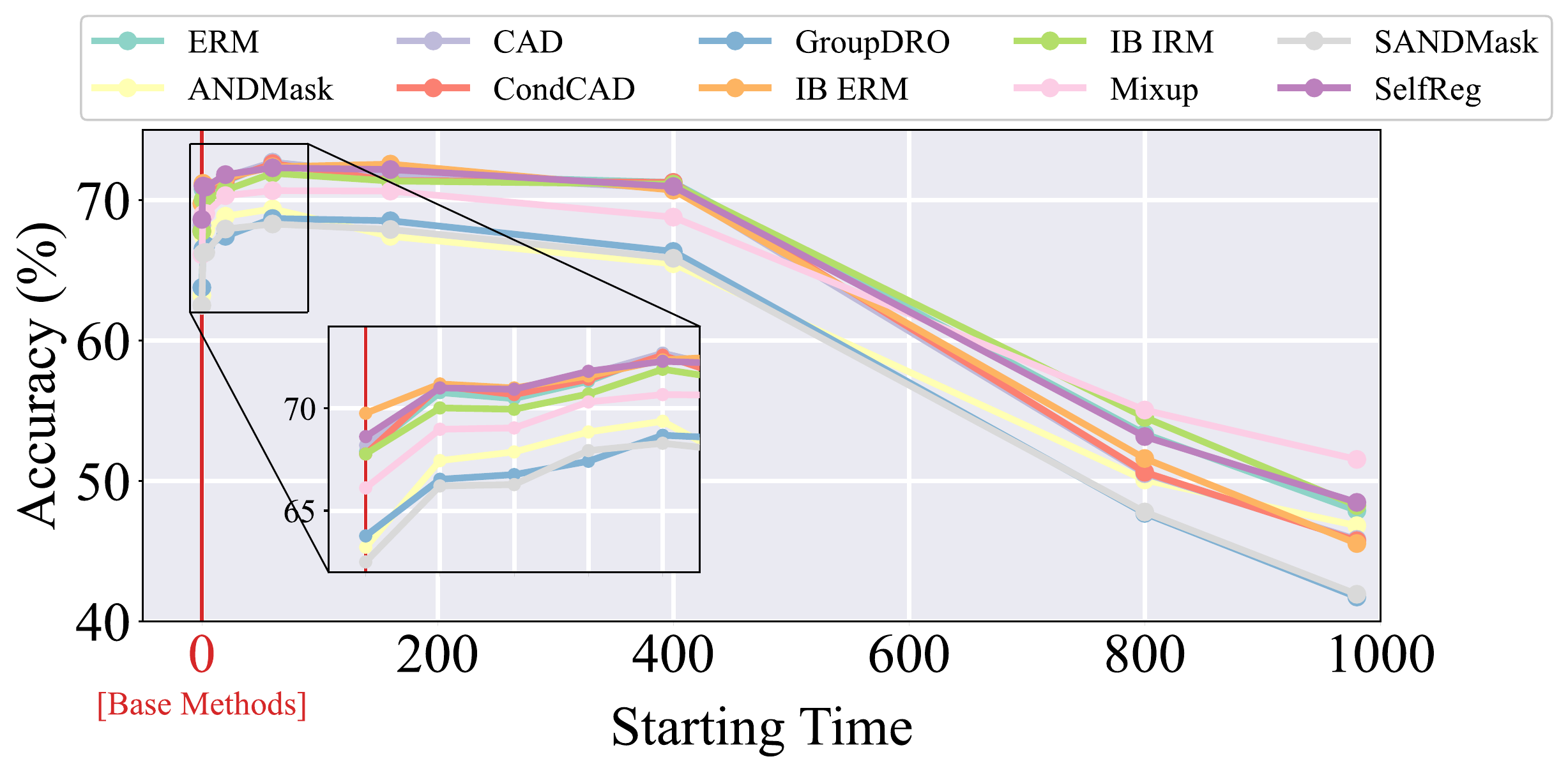}
  \caption{Accuracy v.s. Starting Time $s$ on PACS dataset with the testing domain of ``Cartoon''.}
  \label{fig:proper_s}
  \vspace{-2em}
\end{figure}

%% file: 2figs/hyper.tex
\begin{figure}
  \centering
  \begin{subfigure}{\linewidth}
    \includegraphics[width=\linewidth]{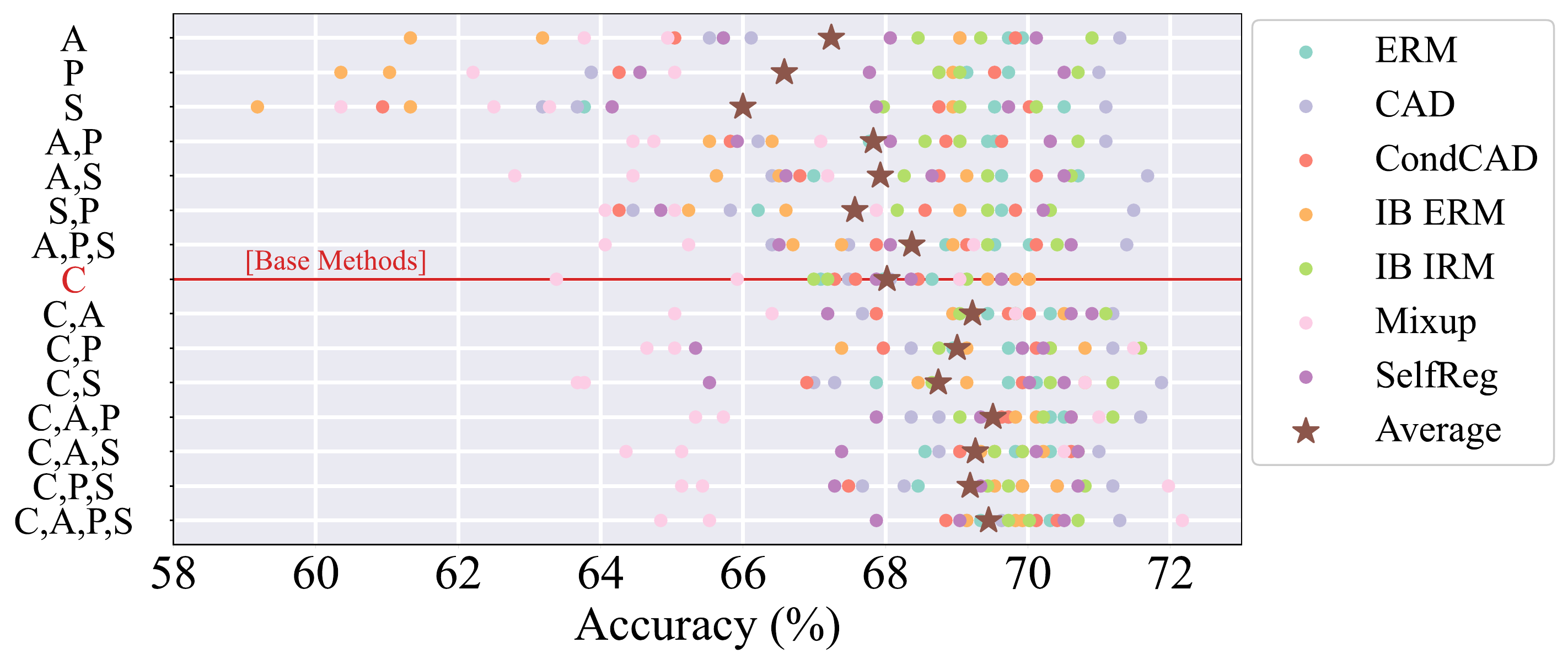}
    \caption{Accuracy v.s. Ensemble Set}
    \label{fig:hyper:a}
  \end{subfigure}\\
  \begin{subfigure}{0.93\linewidth}
    \includegraphics[width=\linewidth]{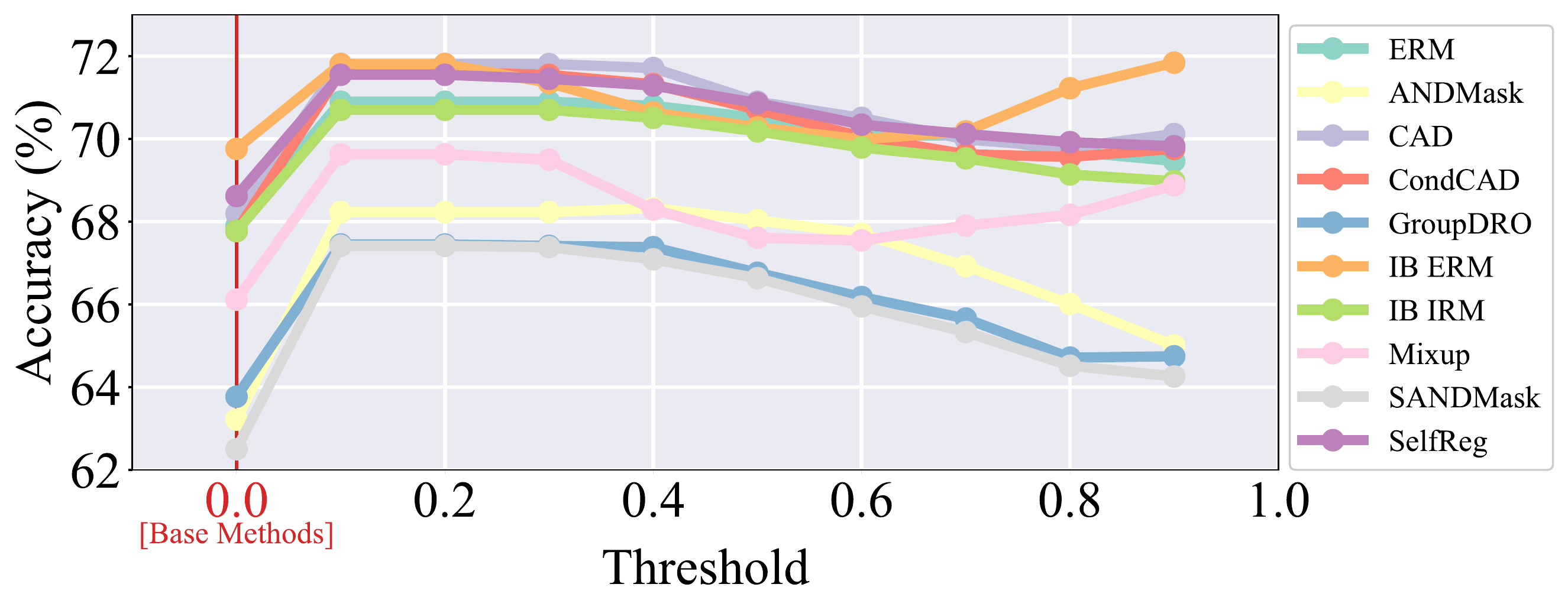}
    \caption{Accuracy v.s. Threshold}
    \label{fig:hyper:b}
  \end{subfigure}
  \vspace{-0.5em}
  \caption{Discussion about the influence of hyperparameters: Ensemble set (\cref{fig:hyper:a}) and Threshold (\cref{fig:hyper:b}) on the PACS dataset with the testing domain of ``Cartoon''. The y-axis label of \cref{fig:hyper:a} indicates the components of the ensemble set with the corresponding accuracy's show on the right. The row of ``\textcolor{myred}{C}'' corresponds to the setting using the original testing images alone. The areas below(above) correspond to the ensembles with(without) the original testing images. }
  \label{fig:hyper}
  \vspace{-1.5em}
\end{figure}

%% file: appendix.tex
\onecolumn
\appendix

\begin{center} 
  \fontsize{18}{20}\selectfont 
  Distribution Shift Inversion for Out-of-Distribution Prediction \\-\textit{Supplementary Material-}
\end{center}

In this document, we provide additional materials that cannot fit into the main manuscript due to the page limit. First, we show experimental results on three more datasets. Extra transferred images are also presented. Next, we provide proof of the Theorem in the main text. 

\section{More Experimental Results and Implementation Details}

\subsection{Performance on ImageNet-R, ImageNet Sketch and CdSprites-5}
The results on ImageNet-R\cite{imagenetr}, ImageNet Sketch\cite{imagenetsketch} and CdSprites-5\cite{cdsprites}  are provided in this section. We use [algorithm]* to indicate the use of \ouralg, the \mycolorbox{green cells} to indicate our method improves the base method, and the \mycolorbox{\textbf{green cells with text in bold}} to indicate our method improves the base method with non-overlapped confidence interval.

\textbf{ImageNet-R} is a variant of ImageNet\cite{imagenet} and contains 30,000 images from 200 classes with different styles. In OoD prediction, combined with a subset of the ImageNet, it is used as a single-training domain classification benchmark. In our experiment, we select the images, whose labels appear in the ImageNet-R, from the training set of ImageNet as  the training set and use ImageNet-R as the testing set. The results is shown in \cref{tab:imagenet-r}. Our method improves the base method by 1.68\% on average.

\textbf{ImageNet Sketch} is another variant of ImageNet and contains 50,000 sketch images collected by Google search engine, which leads to 50 sketch images for each of the ImageNet classes. In the experiments, the original training set of ImageNet is used as the training domain, and ImageNet Sketch is used as the testing domain. The results is shown in \cref{tab:imagenet-r}. Our method improves the base method by 1.38\% on average. This shows that our method is effective for dataset with large number of classes.

\textbf{CdSprites-5} is a variant of DSprites~\cite{dsprites17}. The original DSprites contains white shapes with different scales, rotations, and positions. CdSprites-5 construct a binary classification problem by selecting 2 shapes from DSprites. There are 5 training domains, 1 validation domain and 1 testing domain in CdSprites-5. Instead of using the white shapes, the author color the shape. For different training domain, different color pairs are used. In the validation and testing domain, all colors appear in the training domains are used. Furthermore, spurious correlation is injected to the training set by makes the colors strongly correlated to the shapes. While in the validation and testing set, the colors and the shapes are uncorrelated.  The results is shown in \cref{tab:cdsprites}. Our method achieves significant improvement when combined with base methods. This shows that our method is also able to address the distribution shift induced by covariate shift and spurious correlation. 

\subsection{Transformed Samples}
\cref{fig:sample:cdsprites} shows the training, testing and the transformed samples on the CdSprites dataset. Each testing sample and each transformed sample are corresponded. As shown, in the figure, for almost all sprites, the transformation does not change the shape information, but correct the color information.

\cref{fig:sample:extra} also provide extra transformed sample together with the corresponding OoD samples on PACS, OfficeHome, ImageNet-R and ImageNet Sketch.

\subsection{Diffusion Model Implementation}
For CDSprites-5, a DDPM model is trained from scratch with the default structure in \cite{ddpm}. 
For other datasets, we fine-tune the U-net structure of the stable diffusion~\cite{stabled}\footnote{named sd-v1-4.ckpt in the official github repository.} pretrained on the laion-aesthetics v2 5+. Other important hyperparameters are listed in \cref{table:hyperparametersdm}. AdamW~\cite{adamw} optimizer is used for all the Diffusion Models. After training, the diffusion step is down-scaled to 250 for both types of diffusion model by a linear schedule. 

\subsection{OoD Algorithms Implementation}
 For CDSprites, a 7-layer Convolutional Neural Network with ReLU activation, BatchNorm and residual shortcut is used. For other datasets, EfficientNet is used. All of the algorithms are optimized by Adam~\cite{adam}. To be consistent with previous works, we reuse the  hyperparameter search regions in \cite{domain-bed} and the subsequent updates of its implementation. For completeness, here, we also list the hyperparameter search regions in \cref{table:hyperparameters}.

 \begin{table}
  \begin{minipage}[b]{0.49\textwidth} 
  \centering 
    \input{3tab/cdsprites}
  \end{minipage}
  \begin{minipage}[b]{0.5\textwidth} 
        \centering 
      \begin{minipage}[b]{\textwidth} 
      \centering 
        \input{3tab/imagenetr}
      \end{minipage} \\
      \begin{minipage}[b]{\textwidth} 
      \centering 
     \input{3tab/dmhyper}

      \end{minipage} 
  \end{minipage} 
\end{table}

\input{2figs/cdsprites_samples}
\input{2figs/extra_samples2}

\input{3tab/hyper}

\subsection{No Guidance}
Though guidance is a frequently and widely used guarantee for image generation quality when implementing Diffusion Model, such as the classifier-based~\cite{dmbeatsgan}, the CLIP-based~\cite{Blended,GLIDE}, the classifier-free~\cite{HTC}, and the reference image-based~\cite{ilvr,b2s,Blended} guidance. We free \ouralg\  from using this type of technique based on the following reasons. First, in the OoD prediction task, though can be guessed in a self-supervised learning style, no explicit prompts or labels are available before the OoD prediction is made. This makes the first three guidance techniques inapplicable. Second, the reference image-based guidance is task-specific and strictly preserves certain low-level components of the reference, for example, the low-frequency components~\cite{b2s,ilvr} or parts of the reference~\cite{Blended}. While taking the OoD sample itself as the reference image is possible, there are no low-level components that have a guarantee on the OoD prediction and should be preserved.

\input{2figs/1-D_example}

\subsection{Different Confidence Scores}
We evaluate our method with another two confidence metrics: Maximal Likelihood~\cite{commonloss} and KL-Divergence~\cite{confidence_score_baseline}. The results are shown in \cref{tab:conf_score}. Despite the confidence score used, our method can improve the baseline method on average. The results also indicate the importance of the choice of confidence score. With an improper confidence score, the performance of our method degenerates.
\input{3tab/confidence_score}

\subsection{Experiments Under Domainbed}
Experimental results using Domainbed implementation are shown in \cref{tab:domainbed}. The main difference between the experiment in this subsection and the experiments in the main text and above is the architecture of the predictor. Our method improves the baseline method on average. 
\input{3tab/domainbed}

\subsection{Cost Analyses and Practical Suggestions}
\textbf{Inference}.  Given starting time series $\{s_l\}_{l=0}^L$ (defined in Alg.~1) and $M$ source domains, during inference, the neural network of the base method and each diffusion model forward at most $M\times L+1$ times and $\sum_{l=0}^L s_l$ times, respectively. With diffusion acceleration methods, our method can use smaller $s_l$'s and $L$ to reduce the inference cost.

\textbf{Training}. The training cost of our method is mainly influenced by the number of source domains and the number of samples required.
Our method uses one diffusion model per source domain and one classifier for all source domains. As the number of source-domain grows, the network training cost increases linearly. To reduce the number of diffusion models, one possible way is to group similar domains and train a shared diffusion model for each group. 
Our method uses pre-trained diffusion models and then fine-tunes them to improve sample efficiency. In our experiments, with about $1k$ samples for each source domain, the fine-tuned diffusion models can have desired performance. Few-shot fine-tuning techniques can further reduce the training sample consumption.

\section{Further Discussion on 1-D UDT}

We dig deeper into our 1-D UDT example. to show that the noise space alignment is crucial. The results are plotted in \cref{fig:1dexp}. When the OoD samples are entirely aligned to the noise, label information is lost completely.(\cref{fig:1dexp:T}) Without using noise space alignment, the OoD sample are also Out-of-Distribution with respect to the Diffusion Model. When the original OoD samples are directly fed into the Diffusion Model, the evolution behavior is uncontrollable.(\cref{fig:1dexp:0:m})

\clearpage
\section{Proof of Theorem 1}
\setcounter{theorem}{0}
\begin{theorem}
Given a diffusion model trained on the source distribution $p(\bm{x})$,let $p_t$ denote the distribution at time $t$ in the forward transformation, let $\bar{p}(\bm{x})$ denote the output distribution when the input of the backward process is standard Gaussian noise $\bm{\epsilon}$ whose distribution is denoted by $\rho(\bm{x})$, let $\omega(\bm{x})$ denote the output distribution when the input of backward process is a convex combination $\hat{\bm{X}} = (1-\alpha)\bm{X}'+\alpha \bm{\epsilon}$, where random variable $\bm{X}'$ is sampled following the target distribution $q(\bm{x})$ and $\alpha\in(0,1)$. Under some regularity conditions listed below, we have
\begin{align}
    KL(p||\omega) \le \mathcal{J}_{SM} + KL(p_T||\rho) + \mathcal{F}(\alpha)
\end{align}
\end{theorem}

To prove Theorem 1, we make the following assumptions. Assumptions (i) to (xii) are required for implementing Theorem 1 in \cite{scoreflow}. 
Specifically, 
assumptions (i) \& (ii) require the source distribution and noise distribution to be differentiable and have finite variance, 
assumptions (iii)-(iv) require $f$ or the difference of $f$ in Eq. (2) to be bounded corresponding to its input or the difference of its inputs, 
assumption (iv) requires $g$ in Eq. (2) to be non-zero, 
assumption (vi) requires $p_t$ (defined in Section 3) and its derivative to be bounded. 
assumptions (vii)-(viii) require the score function of $p_t$ or the difference of it to be bounded corresponding to its input or the difference of its inputs, 
assumptions (ix)-(x) require that the estimated score function or the difference of it to be bounded corresponding to its input or the difference of its inputs, 
assumption (x) requires the estimation error is not infinitely large, 
assumption (xii) requires the value of $\bm{X}$ to be bounded, \emph{e.g.},
bounding the values to $[0,255]$ for images.
Assumption (xiii) is used to constrain that $\mathcal{F}$ has a finite first-order derivative.
\begin{enumerate}[label=(\roman*)]
    \item $p(\bm{X}) \in \mcal{C}^2$ and $\E_{\bm{X} \sim p}\big[\norm{\bm{X}}_2^2\big] < \infty$.
    \item $\omega_T(\bm{X}) \in \mcal{C}^2$ and $\E_{\bm{X} \sim \omega_T}\big[ \norm{\bm{X}}_2^2 \big] < \infty$.
    \item $\forall t\in[0,T]: f(\cdot, t) \in \mcal{C}^1$, $\exists C>0~\forall \bm{X} \in \mbb{R}^D, t\in [0,T]: \norm{f(\bm{X}, t)}_2 \leq C(1 + \norm{\bm{X}}_2)$.
    \item $\exists C>0, \forall \bm{X}, \bm{Y} \in \mbb{R}^D: \norm{f(\bm{X}, t) - f(\bm{Y}, t)}_2 \leq C \norm{\bm{X} - \bm{Y}}_2$.
    \item $g \in \mcal{C}$ and $\forall t\in[0,T], |g(t)| > 0$.
    \item For any open bounded set $\mcal{O}$, $\int_0^T \int_\mcal{O} \norm{p_t(\bm{X})}_2^2 + D g(t)^2 \norm{\nabla_{\bm{X}} p_t(\bm{X})}_2^2 \ud \bm{X} \ud t < \infty$.
    \item $\exists C>0~\forall \bm{X} \in \mbb{R}^D, t \in [0,T]: \norm{\nabla_{\bm{X}} \log p_t(\bm{X})}_2 \leq C(1 + \norm{\bm{X}}_2)$.
    \item $\exists C>0, \forall \bm{X}, \bm{Y} \in \mbb{R}^D: \norm{\nabla_{\bm{X}} \log p_t(\bm{X}) - \nabla_{\bm{Y}} \log p_t(\bm{Y})}_2 \leq C \norm{\bm{X} - \bm{Y}}_2$.
    \item $\exists C>0~\forall \bm{X} \in \mbb{R}^D, t \in [0,T]: \norm{\vs_\vtheta(\bm{X}, t)}_2 \leq C(1 + \norm{\bm{X}}_2)$.
    \item $\exists C>0, \forall \bm{X}, \bm{Y} \in \mbb{R}^D: \norm{\vs_\vtheta(\bm{X}, t) - \vs_\vtheta(\bm{Y}, t)}_2 \leq C \norm{\bm{X} - \bm{Y}}_2$.
    \item Novikov's condition: \newline
    $\E\Big[ \exp\Big( \frac{1}{2}\int_0^T \norm{\nabla_{\bm{X}} \log p_t(\bm{X}) - \vs_\vtheta(\bm{X}, t)}_2^2 \ud t \Big) \Big] < \infty$.
    \item $\forall t \in [0,T]~\exists k > 0: p_t(\bm{X}) = O(e^{-\norm{\bm{X}}_2^k})$ as $\norm{\bm{X}}_2 \to \infty$.
    \item $\exists C_1>0$ and  $C_2>0: |\E_{\bm{X} \sim q}\big[\bm{X}\big]|< C_1$ and $|\E_{\bm{X} \sim p_T}\big[\bm{X}\big]|< C_2$.
\end{enumerate}

\begin{proof}
The proof is composed of two parts. First, we bounded the KL-divergence between the $p_T$ and the convex combination $\hat{X}$. Second, we bound the KL-divergence between the generated distribution $\omega$ and the source distribution $p$ and show the convergence of $\mathcal{F}(\alpha)$.

\textbf{Part 1.} The distribution of $\hat{X}$ is in the form of the following convolution, 
\begin{subequations}
\begin{align}
    \omega_T(\bm{x}) &= \int \frac{1}{\alpha(1-\alpha)}\rho(\frac{\bm{x}-\bm{\tau}}{\alpha})q(\frac{\bm{\tau}}{1-\alpha}) d\bm{\tau} \label{app:prf:thm1:0:1} \\
    &= \int \frac{1}{\alpha(1-\alpha)}\frac{1}{\sqrt{2\pi}}e^{-\frac{||\bm{x}-\bm{\tau}||_2^2}{2\alpha^2}}q(\frac{\bm{\tau}}{1-\alpha}) d\bm{\tau} \label{app:prf:thm1:0:2} \\
    &= \int \frac{1}{\alpha(1-\alpha)}\frac{1}{\sqrt{2\pi}}e^{-\frac{||\bm{x}||_2^2}{2\alpha^2}} e^{- (\frac{||\bm{\tau}||_2^2}{2\alpha^2} - \frac{\bm{x}^T\bm{\tau}}{\alpha^2})}q(\frac{\bm{\tau}}{1-\alpha}) d\bm{\tau} \label{app:prf:thm1:0:3} \\
    &= \frac{1}{\alpha}\rho(\frac{\bm{x}}{\alpha}) \int \frac{1}{1-\alpha} e^{- (\frac{||\bm{\tau}||_2^2}{2\alpha^2} - \frac{\bm{x}^T\bm{\tau}}{\alpha^2})}q(\frac{\bm{\tau}}{1-\alpha}) d\bm{\tau} \label{app:prf:thm1:0:4} \\
    &= \frac{1}{\alpha}\rho(\frac{\bm{x}}{\alpha}) \int  e^{- \frac{1}{2\alpha^2}[(1-\alpha)^2||\bm{\nu}||_2^2 - 2(1-\alpha)\bm{x}^T\bm{\nu}]}q(\bm{\nu}) d\bm{\nu} \label{app:prf:thm1:0:5} \\
    &= \rho(\bm{x}) \int  e^{- \frac{1}{2\alpha^2}[(1-\alpha)^2||\bm{\nu}||_2^2 - 2(1-\alpha)\bm{x}^T\bm{\nu}]}q(\bm{\nu}) d\bm{\nu} \label{app:prf:thm1:0:6}\\
    &= \rho(\bm{x}) \mathcal{H}(\alpha,\bm{x}), \label{app:prf:thm1:0:7}
\end{align}
\end{subequations}
where \cref{app:prf:thm1:0:1} is obtained by the independence of $\bm{X}$ and $\bm{\epsilon}$ and the law of changing of random variable, \cref{app:prf:thm1:0:2} and \cref{app:prf:thm1:0:4} are obtained by the definition of the Gaussian distribution, \cref{app:prf:thm1:0:3} is obtained by decomposing the inner square, \cref{app:prf:thm1:0:5} is obtained by substituting $\bm{\tau} = (1-\alpha)\bm{\nu}$, and \cref{app:prf:thm1:0:6} is obtained by using the law of changing of random variable again.

Then, the KL-divergence between $p_T$ and the convex combination $\omega_T$ can be written as 
\begin{subequations}
    \begin{align}
     KL(p_T||\omega_T) &= \int p_T(\bm{x}) \log \frac{p_T(\bm{x})}{\omega_T(\bm{x})} d \bm{x} \label{app:prf:thm1:1:1}\\
    &= \int p_T(\bm{x}) [\log \frac{p_T(\bm{x})}{\rho(\bm{x})} - \log\mathcal{H}(\alpha,\bm{x})] d \bm{x} \label{app:prf:thm1:1:2} \\
    &= KL(p_T||\rho) - \int p_T(\bm{x})\log\mathcal{H}(\alpha,\bm{x})d\bm{x} \label{app:prf:thm1:1:3}\\
    &= KL(p_T||\rho) +\mathcal{F}(\alpha) \label{app:prf:thm1:1:4}
    \end{align}
\end{subequations}
where \cref{app:prf:thm1:1:1} is the definition of the KL-divergence, \cref{app:prf:thm1:1:2} is obtained from \cref{app:prf:thm1:0:7}.  

\textbf{Part 2.} With assumption (i) to (xii), by implementing the Theorem 1 in \cite{scoreflow}, we have 

\begin{align}
    KL(p||\omega) \le \mathcal{J}_{SM} + KL(p_T||\rho) + \mathcal{F}(\alpha).
\end{align}

Because, $\mathcal{F}(1) = 0$ and $\mathcal{F}'(1) = \mathbb{E}_{\bm{X}\sim q}[X]\mathbb{E}_{\bm{X} \sim p_T}[X]$, then by Taylor expansion, we have 
\begin{align}
    \mathcal{F}(\alpha) = (\alpha - 1) \mathbb{E}_{\bm{X}\sim q}[\bm{X}]\mathbb{E}_{\bm{X} \sim p_T}[\bm{X}] + o((\alpha - 1)^2).
\end{align}
Thus, with assumption (xiii), as $\alpha$ goes to $1$, $\mathcal{F}(\alpha)$ converges to $0$.
\end{proof}

%% file: 3tab/cdsprites.tex
\begin{NiceTabular}{lr@{$\pm$}l}[colortbl-like]
\toprule
Dataset     & \multicolumn{2}{c}{CdSprites}  \\ \midrule
ERM                      &     47.46&      0.16\\
ERM*                     & \u{89.39}&  \u{0.52}\\
ANDMask~\cite{andmask}    &     47.56&      0.16\\
ANDMask*                 & \u{89.88}&  \u{0.18}\\
CAD~\cite{cad}            &     47.62&      0.32\\
CAD*                     &   \g48.21&    \g0.32\\
CondCAD~\cite{cad}        &     47.53&      1.40\\
CondCAD*                 &   \g49.32&    \g1.46\\
GroupDRO~\cite{groupDRO}  &     47.27&      0.00\\
GroupDRO*                & \u{89.91}&  \u{0.05}\\
IB\_ERM~\cite{ibirm}       &     47.46&      0.16\\
IB\_ERM*                  & \u{89.39}&  \u{0.52}\\
IB\_IRM~\cite{ibirm}       &     47.36&      0.08\\
IB\_IRM*                  & \u{89.55}&  \u{0.52}\\
Mixup~\cite{mixup}        &     47.69&      0.33\\
Mixup*                   & \u{89.78}&  \u{0.30}\\
SANDMask~\cite{sandmask}  &     47.72&      0.28\\
SANDMask*                & \u{89.55}&  \u{0.44}\\
SelfReg~\cite{selfreg}    &     47.36&      0.08\\
SelfReg*                 & \u{89.52}&  \u{0.52}\\\midrule
Average Gain                &    33.95&    16.38   \\ 
\bottomrule
\end{NiceTabular}
  \caption{The average accuracy $\pm$ the standard deviation of base algorithms w/o our method on CdSprites. The performance is generally boosted when our method is plugged in, whichever base algorithm used.}
  \label{tab:cdsprites}

%% file: 3tab/imagenetr.tex
\begin{NiceTabular}{lr@{$\pm$}lcr@{$\pm$}l}[colortbl-like]
\toprule
Dataset         & \multicolumn{2}{c}{ImageNet-R} & \multicolumn{3}{c}{ImageNet Sketch} \\ \midrule
ERM                 &     36.19&      1.66&  &     20.14&      1.20\\
ERM*                &   \g37.28&    \g0.14&\g&   \g21.08&    \g0.15\\
CORAL               &     36.86&      1.65&  &     18.11&      1.09\\
CORAL*              &   \g38.02&    \g0.06&\g& \u{19.86}&  \u{0.18}\\
RSC                 &     35.58&      1.53&  &     18.44&      0.96\\
RSC*                & \u{37.80}&  \u{0.12}&\g& \u{20.36}&  \u{0.44}\\
SagNet              &     35.26&      0.95&  &     19.50&      1.15\\
SagNet*             & \u{37.49}&  \u{0.67}&\g&   \g20.41&    \g0.16\\ \midrule
Average Gain    &      1.68&      0.55 &   &      1.38&      0.46 \\ \bottomrule
\end{NiceTabular}
  \caption{The average accuracy $\pm$ the standard deviation of base algorithms w/o our method on ImageNet-R and ImageNet Sketch datasets with single training domain setting. The performance is generally boosted when our method is plugged in, whichever base algorithm used.}
  \label{tab:imagenet-r}
  \vspace{1em}

%% file: 3tab/dmhyper.tex
    \begin{tabular}{lll}
        \toprule
        \textbf{Condition} & \textbf{Parameter} & \textbf{Value}\\
        \midrule
        \multirow{3}{*}{DDPM}         & learning rate   & $1e\!-\!4$\\
                                      & batch size      & $32$\\
                                      & diffusion step     & $4,000$\\
        \midrule
        \multirow{3}{*}{Stable Diffusion}   & learning rate   & $1e\!-\!4$\\
                                      & batch size      & $32$\\
                                      & diffusion step  & $1,000$\\
        \bottomrule
    \end{tabular}

    \caption{Diffusion Model Hyperparameters.} 
    \label{table:hyperparametersdm}

%% file: 2figs/cdsprites_samples.tex
\begin{figure*}
  \centering
  \begin{subfigure}{0.32\linewidth}
    \includegraphics[width=\linewidth]{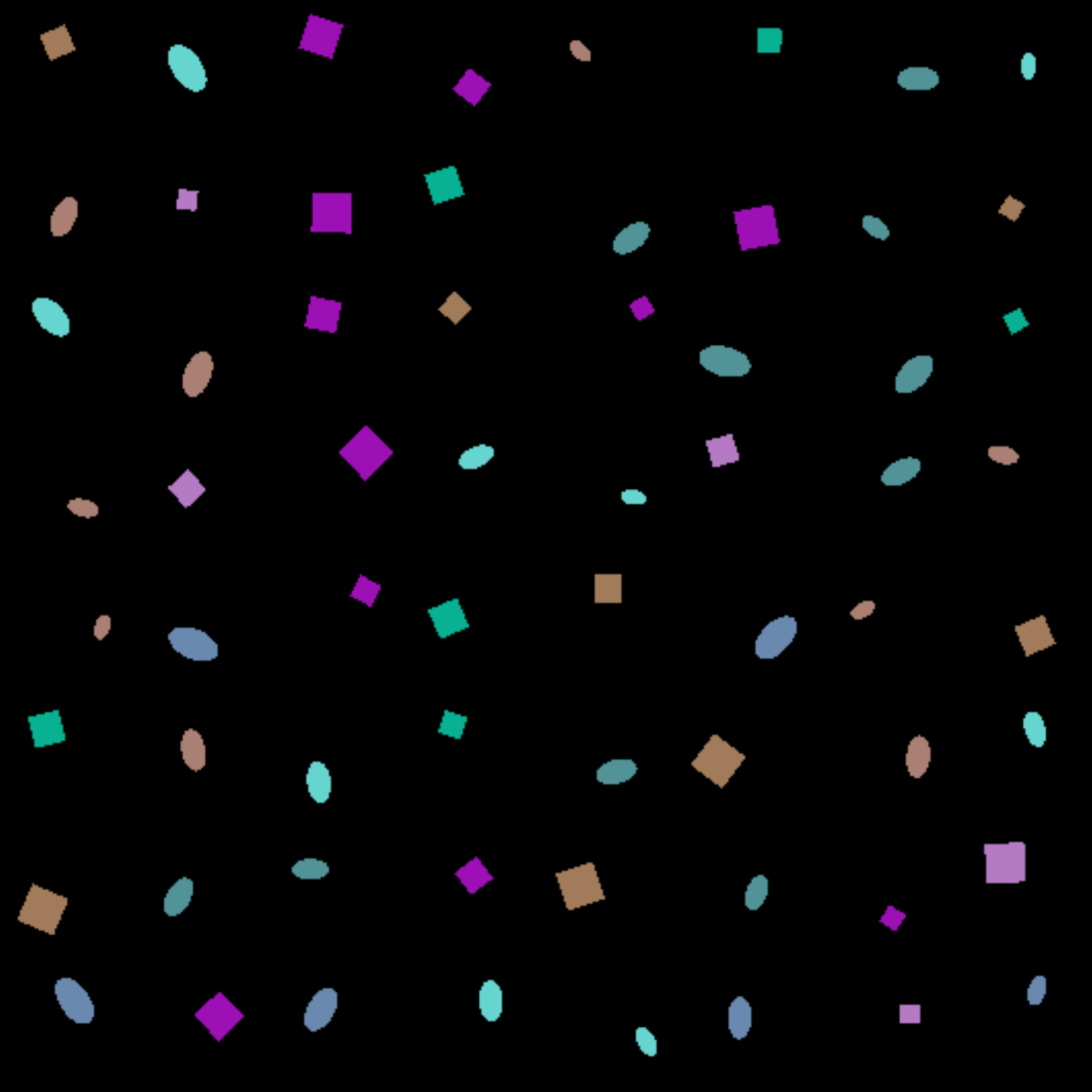}
    \caption{Training Samples}
    \label{fig:sample:cdsprites:a}
  \end{subfigure}
  \begin{subfigure}{0.32\linewidth}
    \includegraphics[width=\linewidth]{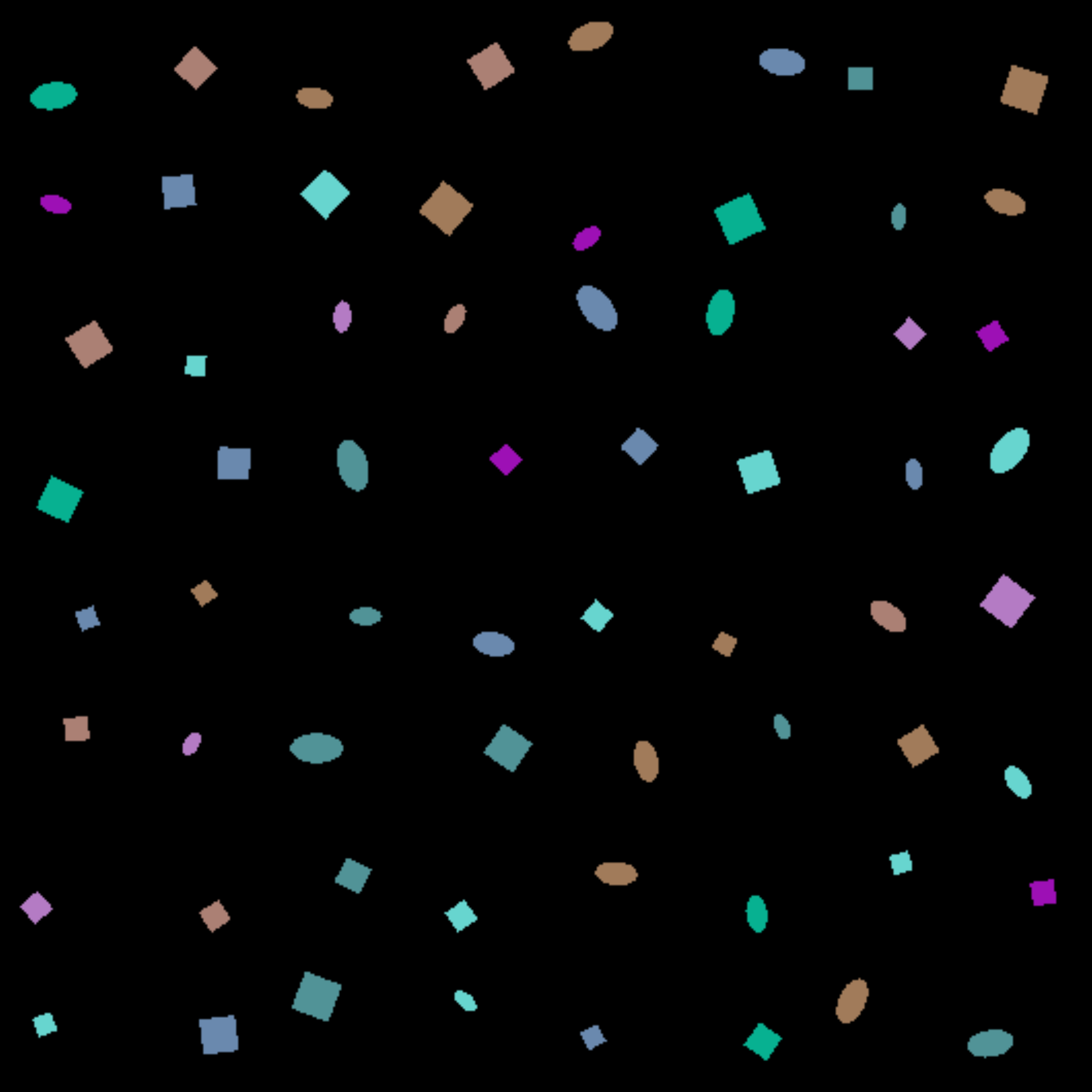}
    \caption{Testing Samples}
    \label{fig:sample:cdsprites:b}
  \end{subfigure}
  \begin{subfigure}{0.32\linewidth}
    \includegraphics[width=\linewidth]{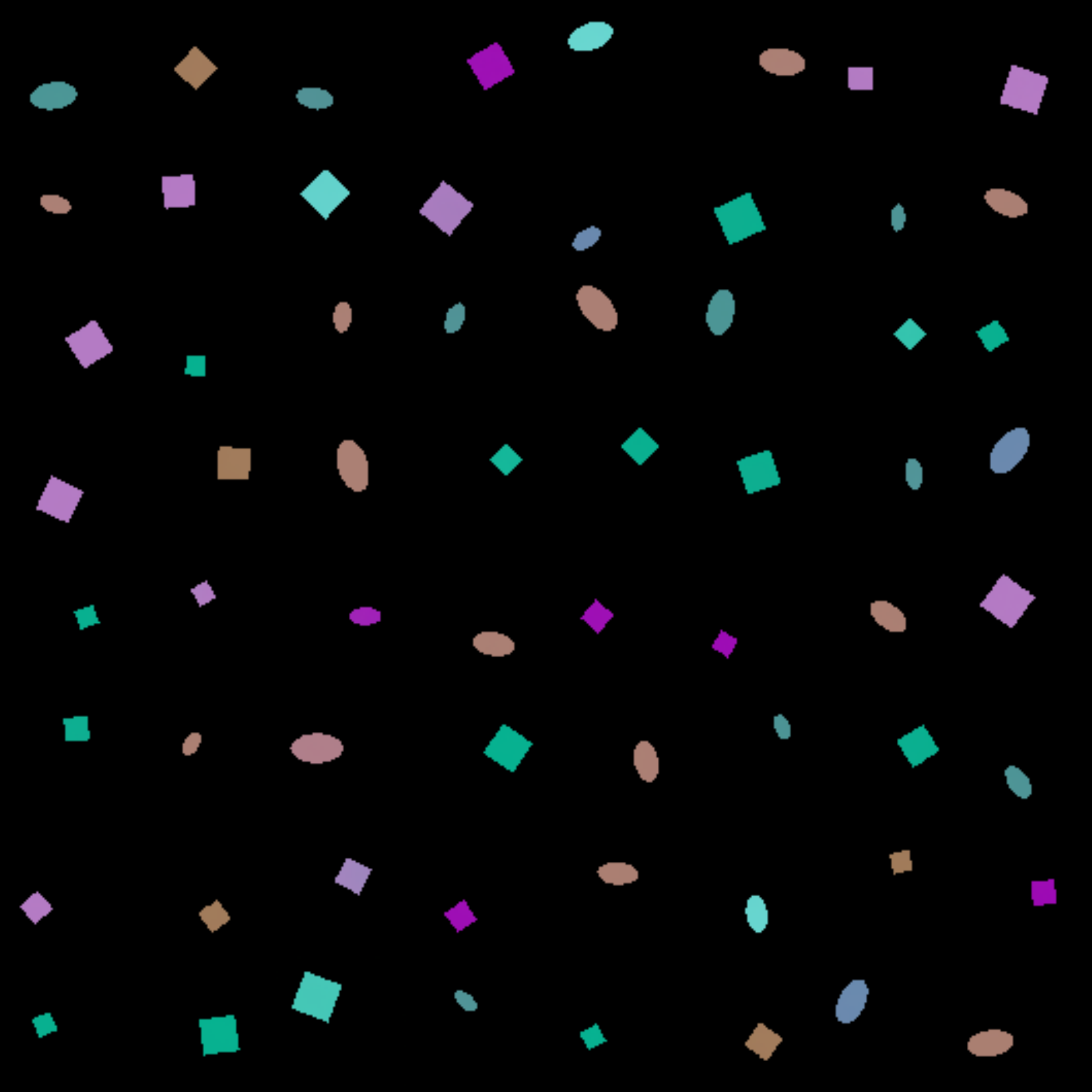}
    \caption{Transformed Samples}
    \label{fig:sample:cdsprites:c}
  \end{subfigure}

  \caption{Samples from CdSprites-5 dataset. Testing and transformed samples are in one-to-one correspondence.}
  \label{fig:sample:cdsprites}
\end{figure*}

%% file: 2figs/extra_samples2.tex
\begin{figure}
  \centering
    \includegraphics[width=\linewidth]{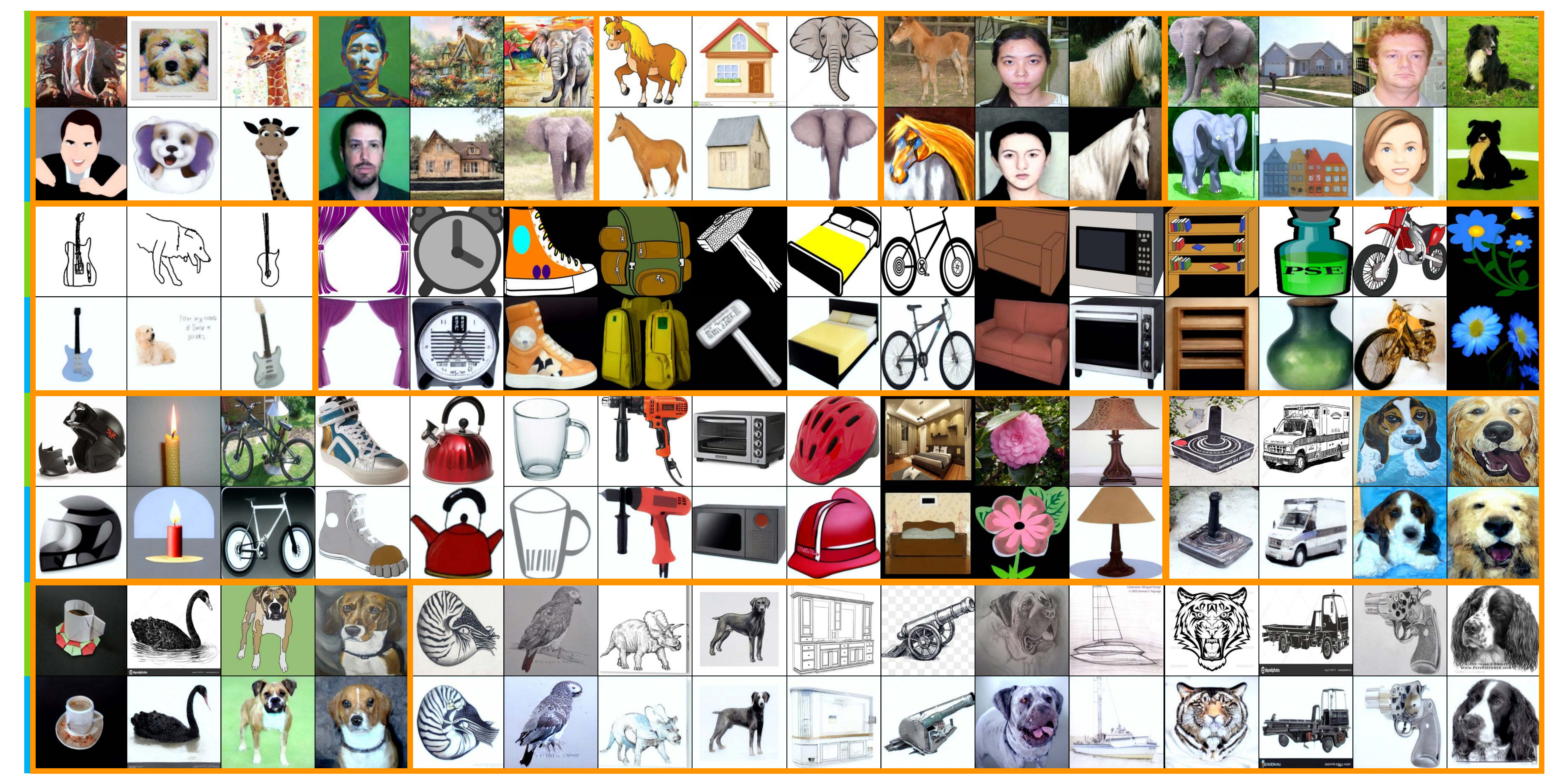}
  \caption{Transformed OoD samples. {\bf \greenrp{Odd rows}} show the original OoD images, and {\bf \bluerp{even rows}} show their transformation results to the source distribution. Samples in the same {\bf \orangerp{box}} are form the same source-target pair.}
  \label{fig:sample:extra}
\end{figure}

%% file: 3tab/hyper.tex
\begin{table*}
    \begin{center}
    { 
    \begin{tabular}{lll}
        \toprule
        \textbf{Condition} & \textbf{Parameter} & \textbf{Region}\\
        \midrule
        \multirow{2}{*}{Common}    & weight decay   & $10^{\text{Uniform}(\!-\!6, \!-\!2)}$\\
                                      & generator weight decay  & $10^{\text{Uniform}(\!-\!6, \!-\!2)}$\\
        \midrule
        \multirow{4}{*}{\makecell[l]{EfficientNet\\ (1000\!-\!class ImageNet)}}       & learning rate  & $10^{\text{Uniform}(\!-\!3.5, \!-\!2.5)}$\\
                                      & batch size      & $2^{\text{Uniform}(8.5, 9.5)}$\\
                                      & generator learning rate  & $10^{\text{Uniform}(\!-\!5, \!-\!3.5)}$\\
                                      & discriminator learning rate  & $10^{\text{Uniform}(\!-\!5, \!-\!3.5)}$\\
        \midrule
        \multirow{4}{*}{\makecell[l]{EfficientNet\\ (Others)}}       & learning rate  & $10^{\text{Uniform}(\!-\!5, \!-\!3.5)}$\\
                                      & batch size      & $2^{\text{Uniform}(3, 5.5)}$\\
                                      & generator learning rate  & $10^{\text{Uniform}(\!-\!5, \!-\!3.5)}$\\
                                      & discriminator learning rate  & $10^{\text{Uniform}(\!-\!5, \!-\!3.5)}$\\
        \midrule
        \multirow{4}{*}{7\!-\!layer CNN}   & learning rate & $10^{\text{Uniform}(\!-\!4.5, \!-\!3.5)}$\\
                                      & batch size    & $2^{\text{Uniform}(3, 9)}$\\
                                      & generator learning rate& $10^{\text{Uniform}(\!-\!4.5, \!-\!2.5)}$\\
                                      & discriminator learning rate & $10^{\text{Uniform}(\!-\!4.5, \!-\!2.5)}$\\
        \midrule
        ANDMask                         & tau  & $\text{Uniform}(0.5, 1)$\\
        \midrule
        \multirow{2}{*}{CAD/ConCAD}          & lambda  & $\text{Choice}(1e\!-\!4, 1e\!-\!3, 1e\!-\!2, 1e\!-\!1, 1, 1e1, 1e2)$\\
                                      & temperature  & $\text{Choice}(0.05, 0.1)$\\
        \midrule
        GroupDRO                     & eta   & $10^{\text{Uniform}(\!-\!1, 1)}$\\
        \midrule
        \multirow{2}{*}{IB ERM}          & lambda  & $10^{\text{Uniform}(\!-\!1, 5)}$\\
                                      & penalty anneal iter  & $int(10^{\text{Uniform}(0, 4)})$\\
        \midrule
        \multirow{4}{*}{IB IRM}          & lambda  & $10^{\text{Uniform}(\!-\!1, 5)}$\\
                                      & penalty anneal iter  & $int(10^{\text{Uniform}(0, 4)})$\\
                                        & irm lambda  & $10^{\text{Uniform}(\!-\!1, 5)}$\\
                                      & irm penalty anneal iter  & $int(10^{\text{Uniform}(0, 4)})$\\
        \midrule
        Mixup                         & alpha  & $10^{\text{Uniform}(0, 4)}$\\
        \midrule
        \multirow{2}{*}{SANDMask}          & tau  & $\text{Uniform}(0.5, 1)$\\
                                      & k  & $10^{\text{Uniform}(\!-\!3, 5)}$\\
        \midrule
        CORAL                     & gamma   & $10^{\text{Uniform}(\!-\!1, 1)}$\\
        \midrule
        \multirow{2}{*}{RSC}          & rsc f drop factor  & $\text{Uniform}(0, 0.5)$\\
                                      & rsc b drop factor  & $\text{Uniform}(0, 0.5)$\\
        \midrule
        SagNet                     & sag w adv   & $10^{\text{Uniform}(\!-\!2, 1)}$\\
        \bottomrule
    \end{tabular}
    }
    \end{center}
      \vspace{-1.5em}
    \caption{OoD Algorithm Hyperparameters.} 
    \label{table:hyperparameters}
      \vspace{-2em}
\end{table*}

%% file: 2figs/1-D_example.tex
\begin{figure}
  \centering
   \begin{subfigure}{0.48\linewidth}
    \includegraphics[width=\linewidth]{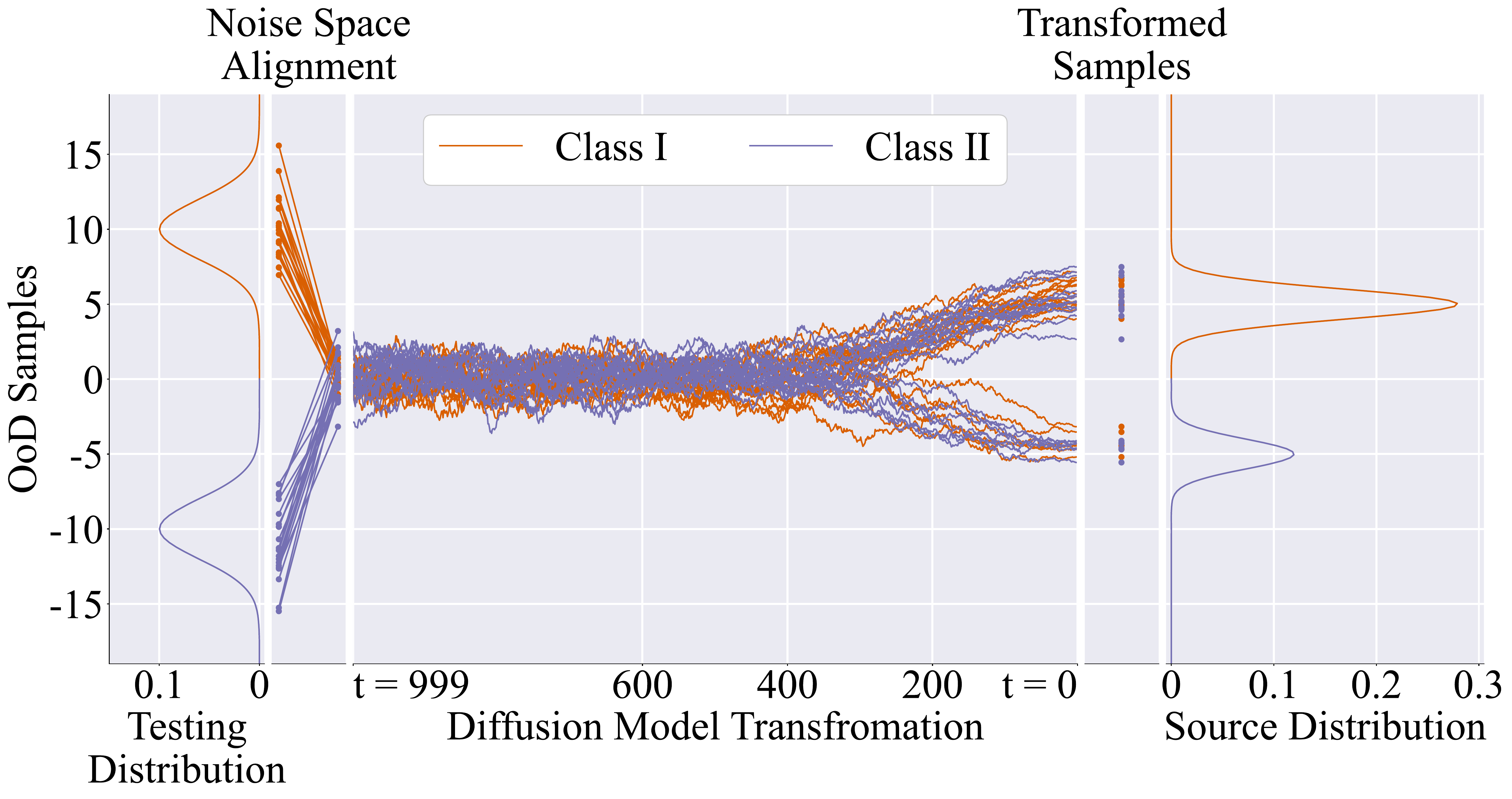}
    \caption{Total Alignment}
    \label{fig:1dexp:T}
  \end{subfigure}
  \begin{subfigure}{0.48\linewidth}
    \includegraphics[width=\linewidth]{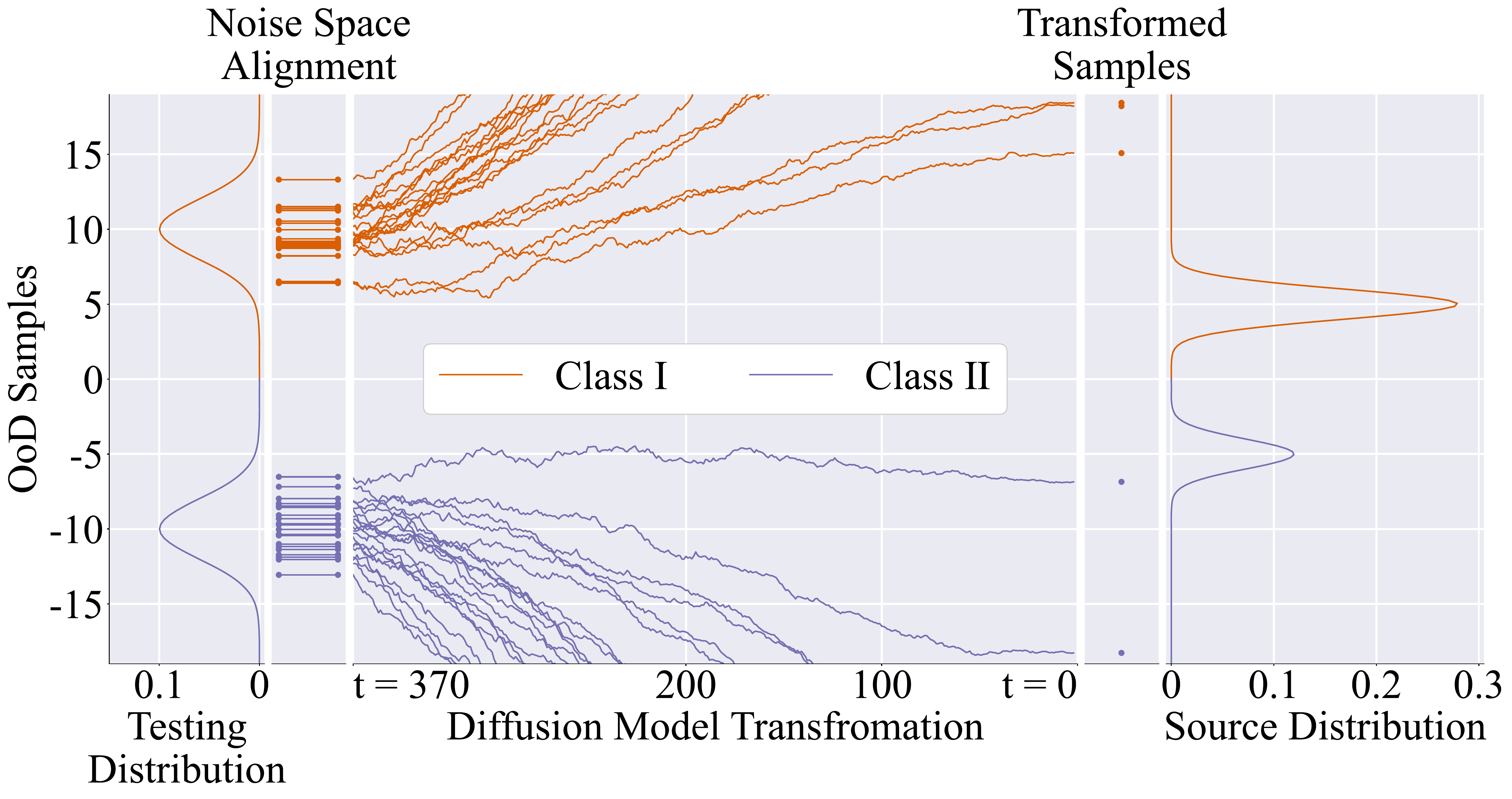}
    \caption{No Alignment, $s = 370$}
    \label{fig:1dexp:0:m}
  \end{subfigure}
  
  \caption{With or without noise space alignment in the 1-D distribution transformation example.}
  \label{fig:1dexp}
\end{figure}

%% file: 3tab/confidence_score.tex
\begin{table}
    \centering
\begin{NiceTabular}{llr@{$\pm$}lr@{$\pm$}lr@{$\pm$}lr@{$\pm$}lr@{$\pm$}l}[colortbl-like]
\toprule
Confidence Score & Algorithm & \multicolumn{2}{c}{A} & \multicolumn{2}{c}{C} & \multicolumn{2}{c}{P} & \multicolumn{2}{c}{S} & \multicolumn{2}{c}{Average}  \\ \midrule
\multirow{4}{*}{Maximal Likelihood} 
& ERM                 &  85.64&      1.17 & 80.44&      0.74 & 96.97&      0.49 & 77.73&      3.97 & 85.20 & 1.60 \\
& ERM*                &  88.96&      0.98 & 85.06&      1.24 & 97.56&      0.21 & 85.42&      3.36 & 89.25 & 1.44 \\
& SelfReg             &  84.83&      2.44 & 78.32&      3.27 & 95.48&      0.62 & 78.65&      4.55 & 84.32 & 2.72 \\
& SelfReg*            &  87.79&      1.91 & 82.65&      2.49 & 96.13&      0.51 & 85.22&      2.80 & 87.95 & 1.93 \\
\midrule
\multirow{4}{*}{KL-Divergence} 
& ERM                   & 85.64&      1.17 & 80.44&      0.74 & 96.97&      0.49 & 77.73&      3.97 & 85.20 &	1.60 \\
& ERM*                  & 85.74&      0.39 & 81.35&      1.60 & 96.35&      0.53 & 80.60&      4.58 & 86.01 &	1.77 \\
& SelfReg               & 84.83&      2.44 & 78.32&      3.27 & 95.48&      0.62 & 78.65&      4.55 & 84.32 &	2.72 \\
& SelfReg*              & 83.79&      3.45 & 79.26&      2.97 & 94.92&      0.55 & 80.63&      4.07 & 84.65 &	2.76 \\
\bottomrule
\end{NiceTabular}
    \vspace{-0.5em}\caption{The average accuracy $\pm$ the standard deviation of base algorithms w/o our method on PACS using different confidence scores.}  \vspace{-0.5em}
  \label{tab:conf_score}
\end{table}

%% file: 3tab/domainbed.tex
\begin{table}
    \centering
\begin{NiceTabular}{llr@{$\pm$}lr@{$\pm$}lr@{$\pm$}lr@{$\pm$}lr@{$\pm$}l}[colortbl-like]
\toprule
Confidence Score & Algorithm & \multicolumn{2}{c}{A} & \multicolumn{2}{c}{C} & \multicolumn{2}{c}{P} & \multicolumn{2}{c}{S} & \multicolumn{2}{c}{Average}  \\ \midrule
\multirow{4}{*}{Training-Set Validation} 
& ERM                 &  84.41&      3.01 & 80.44&      0.74 & 96.81&      0.53 & 79.82&      1.80 & 85.69 & 1.52 \\
& ERM*                &  84.64&      2.42 & 81.38&      1.60 & 96.84&      0.54 & 82.52&      1.04 & 86.91 & 1.40 \\
& SelfReg             &  84.83&      2.44 & 78.55&      4.80 & 95.61&      0.50 & 79.00&      1.21 & 84.39 & 2.24 \\
& SelfReg*            &  84.97&      2.51 & 80.86&      3.69 & 95.61&      0.50 & 80.40&      0.60 & 85.62 & 1.83 \\
\midrule
\multirow{4}{*}{Testing-Set Validation} 
& ERM                   & 85.64&      1.17 & 80.44&      0.74 & 96.42&      0.90 & 77.73&      3.97 & 85.06 &	1.70 \\
& ERM*                  & 85.74&      0.39 & 81.38&      1.60 & 96.88&      0.44 & 80.60&      4.58 & 86.15 &	1.75 \\
& SelfReg               & 84.83&      2.44 & 78.32&      3.27 & 95.64&      0.40 & 78.65&      4.55 & 84.36 &	2.67 \\
& SelfReg*              & 84.97&      2.51 & 79.30&      3.01 & 95.67&      0.33 & 80.53&      4.29 & 85.12 &	2.54 \\
\bottomrule
\end{NiceTabular}
    \vspace{-0.5em}\caption{The average accuracy $\pm$ the standard deviation of base algorithms w/o our method on PACS using Domainbed implementation.}  \vspace{-1em}
  \label{tab:domainbed}
\end{table}

%% file: main.bbl
\begin{thebibliography}{10}\itemsep=-1pt

\bibitem{ibirm}
Kartik Ahuja, Ethan Caballero, Dinghuai Zhang, Yoshua Bengio, Ioannis
  Mitliagkas, and Irina Rish.
\newblock Invariance principle meets information bottleneck for
  out-of-distribution generalization.
\newblock {\em arXiv:2106.06607}, 2021.

\bibitem{irm}
Mart{\'{\i}}n Arjovsky, L{\'{e}}on Bottou, Ishaan Gulrajani, and David
  Lopez{-}Paz.
\newblock Invariant risk minimization.
\newblock {\em arXiv:1907.02893}, 2019.

\bibitem{imgedit2}
Omri Avrahami, Dani Lischinski, and Ohad Fried.
\newblock Blended diffusion for text-driven editing of natural images.
\newblock In {\em CVPR}, 2022.

\bibitem{Blended}
Omri Avrahami, Dani Lischinski, and Ohad Fried.
\newblock Blended diffusion for text-driven editing of natural images.
\newblock In {\em CVPR}, pages 18187--18197, 2022.

\bibitem{decaug}
Haoyue Bai, Rui Sun, Lanqing Hong, Fengwei Zhou, Nanyang Ye, Han{-}Jia Ye,
  S.{-}H.~Gary Chan, and Zhenguo Li.
\newblock Decaug: Out-of-distribution generalization via decomposed feature
  representation and semantic augmentation.
\newblock In {\em AAAI}, 2021.

\bibitem{dmsag}
Dmitry Baranchuk, Andrey Voynov, Ivan Rubachev, Valentin Khrulkov, and Artem
  Babenko.
\newblock Label-efficient semantic segmentation with diffusion models.
\newblock In {\em ICLR}, 2022.

\bibitem{imggen4}
Yaniv Benny and Lior Wolf.
\newblock Dynamic dual-output diffusion models.
\newblock In {\em CVPR}, 2022.

\bibitem{imggen2}
Jooyoung Choi, Sungwon Kim, Yonghyun Jeong, Youngjune Gwon, and Sungroh Yoon.
\newblock {ILVR:} conditioning method for denoising diffusion probabilistic
  models.
\newblock In {\em ICCV}, 2021.

\bibitem{ilvr}
Jooyoung Choi, Sungwon Kim, Yonghyun Jeong, Youngjune Gwon, and Sungroh Yoon.
\newblock {ILVR:} conditioning method for denoising diffusion probabilistic
  models.
\newblock In {\em ICCV}, 2021.

\bibitem{imggen5}
Jooyoung Choi, Jungbeom Lee, Chaehun Shin, Sungwon Kim, Hyunwoo Kim, and
  Sungroh Yoon.
\newblock Perception prioritized training of diffusion models.
\newblock In {\em CVPR}, 2022.

\bibitem{closerfaster}
Hyungjin Chung, Byeongsu Sim, and Jong~Chul Ye.
\newblock Come-closer-diffuse-faster: Accelerating conditional diffusion models
  for inverse problems through stochastic contraction.
\newblock In {\em CVPR}, 2022.

\bibitem{tcp}
Charles Corbi{\`{e}}re, Nicolas Thome, Avner Bar{-}Hen, Matthieu Cord, and
  Patrick P{\'{e}}rez.
\newblock Addressing failure prediction by learning model confidence.
\newblock In {\em NeurIPS}, 2019.

\bibitem{imagenet}
Jia Deng, Wei Dong, Richard Socher, Li{-}Jia Li, Kai Li, and Li Fei{-}Fei.
\newblock Imagenet: {A} large-scale hierarchical image database.
\newblock In {\em CVPR}, 2009.

\bibitem{dmbeatsgan}
Prafulla Dhariwal and Alexander~Quinn Nichol.
\newblock Diffusion models beat gans on image synthesis.
\newblock In {\em NeurIPS}, 2021.

\bibitem{masf}
Qi Dou, Daniel~Coelho de Castro, Konstantinos Kamnitsas, and Ben Glocker.
\newblock Domain generalization via model-agnostic learning of semantic
  features.
\newblock {\em arXiv:1910.13580}, 2019.

\bibitem{mcdropout}
Yarin Gal and Zoubin Ghahramani.
\newblock Dropout as a bayesian approximation: Representing model uncertainty
  in deep learning.
\newblock In {\em ICML}, 2016.

\bibitem{dann}
Yaroslav Ganin, Evgeniya Ustinova, Hana Ajakan, Pascal Germain, Hugo
  Larochelle, Fran{\c{c}}ois Laviolette, Mario Marchand, and Victor Lempitsky.
\newblock Domain-adversarial training of neural networks.
\newblock {\em JMLR}, 2016.

\bibitem{b2s}
Jin Gao, Jialing Zhang, Xihui Liu, Trevor Darrell, Evan Shelhamer, and Dequan
  Wang.
\newblock Back to the source: Diffusion-driven test-time adaptation.
\newblock {\em CoRR}, 2022.

\bibitem{selective_classification}
Yonatan Geifman and Ran El{-}Yaniv.
\newblock Selective classification for deep neural networks.
\newblock In {\em NeurIPS}, 2017.

\bibitem{gan}
Ian~J. Goodfellow, Jean Pouget{-}Abadie, Mehdi Mirza, Bing Xu, David
  Warde{-}Farley, Sherjil Ozair, Aaron~C. Courville, and Yoshua Bengio.
\newblock Generative adversarial nets.
\newblock In {\em NeurIPS}, 2014.

\bibitem{text2imge1}
Shuyang Gu, Dong Chen, Jianmin Bao, Fang Wen, Bo Zhang, Dongdong Chen, Lu Yuan,
  and Baining Guo.
\newblock Vector quantized diffusion model for text-to-image synthesis.
\newblock In {\em CVPR}, 2022.

\bibitem{domain-bed}
Ishaan Gulrajani and David Lopez{-}Paz.
\newblock In search of lost domain generalization.
\newblock In {\em ICLR}, 2021.

\bibitem{cnbb}
Yue He, Zheyan Shen, and Peng Cui.
\newblock Towards non-i.i.d. image classification: {A} dataset and baselines.
\newblock {\em Pattern Recognit.}, 2021.

\bibitem{imagenetr}
Dan Hendrycks, Steven Basart, Norman Mu, Saurav Kadavath, Frank Wang, Evan
  Dorundo, Rahul Desai, Tyler Zhu, Samyak Parajuli, Mike Guo, Dawn Song, Jacob
  Steinhardt, and Justin Gilmer.
\newblock The many faces of robustness: {A} critical analysis of
  out-of-distribution generalization.
\newblock In {\em ICCV}, 2021.

\bibitem{confidence_score_baseline}
Dan Hendrycks and Kevin Gimpel.
\newblock A baseline for detecting misclassified and out-of-distribution
  examples in neural networks.
\newblock In {\em ICLR}, 2017.

\bibitem{ddpm}
Jonathan Ho, Ajay Jain, and Pieter Abbeel.
\newblock Denoising diffusion probabilistic models.
\newblock In {\em NIPS}, 2020.

\bibitem{videogen}
Jonathan Ho, Tim Salimans, Alexey~A. Gritsenko, William Chan, Mohammad Norouzi,
  and David~J. Fleet.
\newblock Video diffusion models.
\newblock {\em CoRR}, 2022.

\bibitem{cycada}
Judy Hoffman, Eric Tzeng, Taesung Park, Jun{-}Yan Zhu, Phillip Isola, Kate
  Saenko, Alexei~A. Efros, and Trevor Darrell.
\newblock Cycada: Cycle-consistent adversarial domain adaptation.
\newblock In {\em ICML}, 2018.

\bibitem{graphgen2}
Emiel Hoogeboom, Victor~Garcia Satorras, Cl{\'{e}}ment Vignac, and Max Welling.
\newblock Equivariant diffusion for molecule generation in 3d.
\newblock In {\em ICML}, 2022.

\bibitem{rsc}
Zeyi Huang, Haohan Wang, Eric~P. Xing, and Dong Huang.
\newblock Self-challenging improves cross-domain generalization.
\newblock In Andrea Vedaldi, Horst Bischof, Thomas Brox, and Jan{-}Michael
  Frahm, editors, {\em AAAI}, 2020.

\bibitem{knnscore}
Heinrich Jiang, Been Kim, Melody~Y. Guan, and Maya~R. Gupta.
\newblock To trust or not to trust {A} classifier.
\newblock In {\em NeurIPS}, 2018.

\bibitem{selfreg}
Daehee Kim, Seunghyun Park, Jinkyu Kim, and Jaekoo Lee.
\newblock Selfreg: Self-supervised contrastive regularization for domain
  generalization.
\newblock {\em arXiv:2104.09841}, 2021.

\bibitem{adam}
Diederik~P. Kingma and Jimmy Ba.
\newblock Adam: {A} method for stochastic optimization.
\newblock In {\em ICLR}, 2015.

\bibitem{vae}
Diederik~P. Kingma and Max Welling.
\newblock Auto-encoding variational bayes.
\newblock In {\em ICLR}, 2014.

\bibitem{timegen}
Zhifeng Kong, Wei Ping, Jiaji Huang, Kexin Zhao, and Bryan Catanzaro.
\newblock Diffwave: {A} versatile diffusion model for audio synthesis.
\newblock In {\em ICLR}, 2021.

\bibitem{iga}
Masanori Koyama and Shoichiro Yamaguchi.
\newblock When is invariance useful in an out-of-distribution generalization
  problem?
\newblock {\em arXiv:2008.01883}, 2021.

\bibitem{commonloss}
Balaji Lakshminarayanan, Alexander Pritzel, and Charles Blundell.
\newblock Simple and scalable predictive uncertainty estimation using deep
  ensembles.
\newblock In {\em NeurIPS}, 2017.

\bibitem{voicegen}
Max W.~Y. Lam, Jun Wang, Dan Su, and Dong Yu.
\newblock {BDDM:} bilateral denoising diffusion models for fast and
  high-quality speech synthesis.
\newblock In {\em ICLR}, 2022.

\bibitem{pacs}
Da Li, Yongxin Yang, Yi{-}Zhe Song, and Timothy~M. Hospedales.
\newblock Deeper, broader and artier domain generalization.
\newblock In {\em ICCV}, 2017.

\bibitem{mldg}
Da Li, Yongxin Yang, Yi{-}Zhe Song, and Timothy~M. Hospedales.
\newblock Learning to generalize: Meta-learning for domain generalization.
\newblock In Sheila~A. McIlraith and Kilian~Q. Weinberger, editors, {\em AAAI},
  2018.

\bibitem{textgen}
Xiang~Lisa Li, John Thickstun, Ishaan Gulrajani, Percy Liang, and Tatsunori~B.
  Hashimoto.
\newblock Diffusion-lm improves controllable text generation.
\newblock {\em CoRR}, 2022.

\bibitem{cdan}
Ya Li, Xinmei Tian, Mingming Gong, Yajing Liu, Tongliang Liu, Kun Zhang, and
  Dacheng Tao.
\newblock Deep domain generalization via conditional invariant adversarial
  networks.
\newblock In {\em ECCV}, 2018.

\bibitem{oodd1}
Shiyu Liang, Yixuan Li, and R. Srikant.
\newblock Enhancing the reliability of out-of-distribution image detection in
  neural networks.
\newblock In {\em ICLR}, 2018.

\bibitem{i2ida}
Ming{-}Yu Liu, Thomas~M. Breuel, and Jan Kautz.
\newblock Unsupervised image-to-image translation networks.
\newblock In {\em NeurIPS}, 2017.

\bibitem{dc}
Songhua Liu, Kai Wang, Xingyi Yang, Jingwen Ye, and Xinchao Wang.
\newblock Dataset distillation via factorization.
\newblock {\em NeurIPS}, 2022.

\bibitem{detectenergy}
Weitang Liu, Xiaoyun Wang, John~D. Owens, and Yixuan Li.
\newblock Energy-based out-of-distribution detection.
\newblock In {\em NIPS}, 2020.

\bibitem{adamw}
Ilya Loshchilov and Frank Hutter.
\newblock Decoupled weight decay regularization.
\newblock In {\em ICLR}, 2019.

\bibitem{imginpt}
Andreas Lugmayr, Martin Danelljan, Andr{\'{e}}s Romero, Fisher Yu, Radu
  Timofte, and Luc~Van Gool.
\newblock Repaint: Inpainting using denoising diffusion probabilistic models.
\newblock In {\em CVPR}, 2022.

\bibitem{pointcloudgen2}
Shitong Luo and Wei Hu.
\newblock Diffusion probabilistic models for 3d point cloud generation.
\newblock In {\em CVPR}, 2021.

\bibitem{pointcloudgen3}
Zhaoyang Lyu, Zhifeng Kong, Xudong Xu, Liang Pan, and Dahua Lin.
\newblock A conditional point diffusion-refinement paradigm for 3d point cloud
  completion.
\newblock In {\em ICLR}, 2022.

\bibitem{dsprites17}
Loic Matthey, Irina Higgins, Demis Hassabis, and Alexander Lerchner.
\newblock dsprites: Disentanglement testing sprites dataset.
\newblock https://github.com/deepmind/dsprites-dataset/, 2017.

\bibitem{imgedit1}
Chenlin Meng, Yutong He, Yang Song, Jiaming Song, Jiajun Wu, Jun{-}Yan Zhu, and
  Stefano Ermon.
\newblock Sdedit: Guided image synthesis and editing with stochastic
  differential equations.
\newblock In {\em ICLR}, 2022.

\bibitem{SDEdit}
Chenlin Meng, Yutong He, Yang Song, Jiaming Song, Jiajun Wu, Jun{-}Yan Zhu, and
  Stefano Ermon.
\newblock Sdedit: Guided image synthesis and editing with stochastic
  differential equations.
\newblock In {\em ICLR}, 2022.

\bibitem{typicality}
Eric~T. Nalisnick, Akihiro Matsukawa, Yee~Whye Teh, and Balaji
  Lakshminarayanan.
\newblock Detecting out-of-distribution inputs to deep generative models using
  a test for typicality.
\newblock {\em CoRR}, 2019.

\bibitem{sagnet}
Hyeonseob Nam, HyunJae Lee, Jongchan Park, Wonjun Yoon, and Donggeun Yoo.
\newblock Reducing domain gap by reducing style bias.
\newblock In {\em CVPR}, 2021.

\bibitem{text2imge3}
Alexander~Quinn Nichol, Prafulla Dhariwal, Aditya Ramesh, Pranav Shyam, Pamela
  Mishkin, Bob McGrew, Ilya Sutskever, and Mark Chen.
\newblock {GLIDE:} towards photorealistic image generation and editing with
  text-guided diffusion models.
\newblock In {\em ICML}, 2022.

\bibitem{GLIDE}
Alexander~Quinn Nichol, Prafulla Dhariwal, Aditya Ramesh, Pranav Shyam, Pamela
  Mishkin, Bob McGrew, Ilya Sutskever, and Mark Chen.
\newblock {GLIDE:} towards photorealistic image generation and editing with
  text-guided diffusion models.
\newblock In {\em ICML}, 2022.

\bibitem{dm4ad}
Weili Nie, Brandon Guo, Yujia Huang, Chaowei Xiao, Arash Vahdat, and Animashree
  Anandkumar.
\newblock Diffusion models for adversarial purification.
\newblock In {\em ICML}, 2022.

\bibitem{andmask}
Giambattista Parascandolo, Alexander Neitz, Antonio Orvieto, Luigi Gresele, and
  Bernhard Sch{\"{o}}lkopf.
\newblock Learning explanations that are hard to vary.
\newblock {\em arXiv:2009.00329}, 2020.

\bibitem{icp}
Jonas Peters, Peter Bühlmann, and Nicolai Meinshausen.
\newblock Causal inference by using invariant prediction: identification and
  confidence intervals.
\newblock {\em Journal of the Royal Statistical Society: Series B (Statistical
  Methodology)}, 78(5):947--1012, 2016.

\bibitem{sd}
Mohammad Pezeshki, S{\'{e}}kou{-}Oumar Kaba, Yoshua Bengio, Aaron~C. Courville,
  Doina Precup, and Guillaume Lajoie.
\newblock Gradient starvation: {A} learning proclivity in neural networks.
\newblock {\em arXiv:2011.09468}, 2020.

\bibitem{text2speech}
Vadim Popov, Ivan Vovk, Vladimir Gogoryan, Tasnima Sadekova, and Mikhail~A.
  Kudinov.
\newblock Grad-tts: {A} diffusion probabilistic model for text-to-speech.
\newblock In {\em ICML}, 2021.

\bibitem{imggen1}
Konpat Preechakul, Nattanat Chatthee, Suttisak Wizadwongsa, and Supasorn
  Suwajanakorn.
\newblock Diffusion autoencoders: Toward a meaningful and decodable
  representation.
\newblock In {\em CVPR}, 2022.

\bibitem{text2imge2}
Aditya Ramesh, Prafulla Dhariwal, Alex Nichol, Casey Chu, and Mark Chen.
\newblock Hierarchical text-conditional image generation with {CLIP} latents.
\newblock {\em CoRR}, 2022.

\bibitem{HTC}
Aditya Ramesh, Prafulla Dhariwal, Alex Nichol, Casey Chu, and Mark Chen.
\newblock Hierarchical text-conditional image generation with {CLIP} latents.
\newblock {\em CoRR}, 2022.

\bibitem{timepred}
Kashif Rasul, Calvin Seward, Ingmar Schuster, and Roland Vollgraf.
\newblock Autoregressive denoising diffusion models for multivariate
  probabilistic time series forecasting.
\newblock In {\em ICML}, 2021.

\bibitem{llr}
Jie Ren, Peter~J. Liu, Emily Fertig, Jasper Snoek, Ryan Poplin, Mark~A.
  DePristo, Joshua~V. Dillon, and Balaji Lakshminarayanan.
\newblock Likelihood ratios for out-of-distribution detection.
\newblock In {\em NeurIPS}, 2019.

\bibitem{stabled}
Robin Rombach, Andreas Blattmann, Dominik Lorenz, Patrick Esser, and
  Bj{\"{o}}rn Ommer.
\newblock High-resolution image synthesis with latent diffusion models.
\newblock In {\em CVPR}, 2022.

\bibitem{cad}
Yangjun Ruan, Yann Dubois, and Chris~J. Maddison.
\newblock Optimal representations for covariate shift.
\newblock In {\em ICLR}, 2022.

\bibitem{groupDRO}
Shiori Sagawa, Pang~Wei Koh, Tatsunori~B. Hashimoto, and Percy Liang.
\newblock Distributionally robust neural networks for group shifts: On the
  importance of regularization for worst-case generalization.
\newblock {\em arXiv:1911.08731}, 2019.

\bibitem{text2imge4}
Chitwan Saharia, William Chan, Saurabh Saxena, Lala Li, Jay Whang, Emily
  Denton, Seyed Kamyar~Seyed Ghasemipour, Burcu~Karagol Ayan, S.~Sara Mahdavi,
  Rapha~Gontijo Lopes, Tim Salimans, Jonathan Ho, David~J. Fleet, and Mohammad
  Norouzi.
\newblock Photorealistic text-to-image diffusion models with deep language
  understanding.
\newblock {\em CoRR}, 2022.

\bibitem{imggen3}
Vikash Sehwag, Caner Hazirbas, Albert Gordo, Firat Ozgenel, and Cristian
  Canton{-}Ferrer.
\newblock Generating high fidelity data from low-density regions using
  diffusion models.
\newblock In {\em CVPR}, 2022.

\bibitem{complexity}
Joan Serr{\`{a}}, David {\'{A}}lvarez, Vicen{\c{c}} G{\'{o}}mez, Olga
  Slizovskaia, Jos{\'{e}}~F. N{\'{u}}{\~{n}}ez, and Jordi Luque.
\newblock Input complexity and out-of-distribution detection with
  likelihood-based generative models.
\newblock In {\em ICLR}, 2020.

\bibitem{sandmask}
Soroosh Shahtalebi, Jean{-}Christophe Gagnon{-}Audet, Touraj Laleh, Mojtaba
  Faramarzi, Kartik Ahuja, and Irina Rish.
\newblock Sand-mask: An enhanced gradient masking strategy for the discovery of
  invariances in domain generalization.
\newblock {\em arXiv:2106.02266}, 2021.

\bibitem{ood-survey}
Zheyan Shen, Jiashuo Liu, Yue He, Xingxuan Zhang, Renzhe Xu, Han Yu, and Peng
  Cui.
\newblock Towards out-of-distribution generalization: {A} survey.
\newblock {\em CoRR}, 2021.

\bibitem{cdsprites}
Yuge Shi, Jeffrey Seely, Philip H.~S. Torr, Siddharth Narayanaswamy, Awni~Y.
  Hannun, Nicolas Usunier, and Gabriel Synnaeve.
\newblock Gradient matching for domain generalization.
\newblock In {\em ICLR}, 2022.

\bibitem{ddim}
Jiaming Song, Chenlin Meng, and Stefano Ermon.
\newblock Denoising diffusion implicit models.
\newblock In {\em ICLR}, 2021.

\bibitem{scoreflow}
Yang Song, Conor Durkan, Iain Murray, and Stefano Ermon.
\newblock Maximum likelihood training of score-based diffusion models.
\newblock In {\em NIPS}, 2021.

\bibitem{coral}
Baochen Sun, Jiashi Feng, and Kate Saenko.
\newblock Return of frustratingly easy domain adaptation.
\newblock In Dale Schuurmans and Michael~P. Wellman, editors, {\em AAAI}, 2016.

\bibitem{timeimp}
Yusuke Tashiro, Jiaming Song, Yang Song, and Stefano Ermon.
\newblock {CSDI:} conditional score-based diffusion models for probabilistic
  time series imputation.
\newblock In {\em NeurIPS}, 2021.

\bibitem{officehome}
Hemanth Venkateswara, Jose Eusebio, Shayok Chakraborty, and Sethuraman
  Panchanathan.
\newblock Deep hashing network for unsupervised domain adaptation.
\newblock In {\em CVPR}, 2017.

\bibitem{calibration}
Yoav Wald, Amir Feder, Daniel Greenfeld, and Uri Shalit.
\newblock On calibration and out-of-domain generalization.
\newblock In {\em NeurIPS}, 2021.

\bibitem{imagenetsketch}
Haohan Wang, Songwei Ge, Zachary Lipton, and Eric~P Xing.
\newblock Learning robust global representations by penalizing local predictive
  power.
\newblock In {\em NeurIPS}, 2019.

\bibitem{ood-survey2}
Jindong Wang, Cuiling Lan, Chang Liu, Yidong Ouyang, and Tao Qin.
\newblock Generalizing to unseen domains: {A} survey on domain generalization.
\newblock In {\em IJCAI}, 2021.

\bibitem{trilemma}
Zhisheng Xiao, Karsten Kreis, and Arash Vahdat.
\newblock Tackling the generative learning trilemma with denoising diffusion
  gans.
\newblock In {\em ICLR}, 2022.

\bibitem{graphgen1}
Minkai Xu, Lantao Yu, Yang Song, Chence Shi, Stefano Ermon, and Jian Tang.
\newblock Geodiff: {A} geometric diffusion model for molecular conformation
  generation.
\newblock In {\em ICLR}, 2022.

\bibitem{mixup}
Shen Yan, Huan Song, Nanxiang Li, Lincan Zou, and Liu Ren.
\newblock Improve unsupervised domain adaptation with mixup training.
\newblock {\em arXiv:2001.00677}, 2020.

\bibitem{nreuse2}
Xingyi Yang, Jingwen Ye, and Xinchao Wang.
\newblock Factorizing knowledge in neural networks.
\newblock In {\em ECCV}, 2022.

\bibitem{nreuse3}
Xingyi Yang, Daquan Zhou, Jiashi Feng, and Xinchao Wang.
\newblock Diffusion probabilistic model made slim.
\newblock {\em CVPR}, 2023.

\bibitem{nreuse1}
Xingyi Yang, Daquan Zhou, Songhua Liu, Jingwen Ye, and Xinchao Wang.
\newblock Deep model reassembly.
\newblock {\em NeurIPS}, 2022.

\bibitem{nreuse4}
Jingwen Ye, Songhua Liu, and Xinchao Wang.
\newblock Partial network cloning.
\newblock {\em CVPR}, 2023.

\bibitem{ood-bench}
Nanyang Ye, Kaican Li, Lanqing Hong, Haoyue Bai, Yiting Chen, Fengwei Zhou, and
  Zhenguo Li.
\newblock Ood-bench: Benchmarking and understanding out-of-distribution
  generalization datasets and algorithms.
\newblock {\em arXiv:2106.03721}, 2021.

\bibitem{textmodeling}
Peiyu Yu, Sirui Xie, Xiaojian Ma, Baoxiong Jia, Bo Pang, Ruiqi Gao, Yixin Zhu,
  Song{-}Chun Zhu, and Ying~Nian Wu.
\newblock Latent diffusion energy-based model for interpretable text modelling.
\newblock In {\em ICML}, 2022.

\bibitem{subnet}
Dinghuai Zhang, Kartik Ahuja, Yilun Xu, Yisen Wang, and Aaron~C. Courville.
\newblock Can subnetwork structure be the key to out-of-distribution
  generalization?
\newblock In {\em ICML}, 2021.

\bibitem{segmentationda}
Sicheng Zhao, Bo Li, Xiangyu Yue, Yang Gu, Pengfei Xu, Runbo Hu, Hua Chai, and
  Kurt Keutzer.
\newblock Multi-source domain adaptation for semantic segmentation.
\newblock In {\em NeurIPS}, 2019.

\bibitem{pointcloudgen1}
Linqi Zhou, Yilun Du, and Jiajun Wu.
\newblock 3d shape generation and completion through point-voxel diffusion.
\newblock In {\em ICCV}, 2021.

\end{thebibliography}
